\documentclass[11pt]{article}

\usepackage{fullpage}

\usepackage{url}

\usepackage[english]{babel}
\usepackage[utf8]{inputenc}
\usepackage{xspace}
\usepackage{amsmath,amsthm,amssymb}
\usepackage{lmodern}
\usepackage{courier}
\usepackage{listings}
\usepackage[algo2e,ruled,vlined,linesnumbered]{algorithm2e}
\newcommand{\assign}{\leftarrow}

\usepackage{xcolor}
\usepackage{tikz}
\usepackage{graphicx}
\usepackage{textcomp}
\usepackage{lmodern}

\allowdisplaybreaks[4]
\newcommand{\tool}{\textsc{{IOHprofiler}}\xspace}
\newcommand{\ioh}{iterative optimization heuristic\xspace}

\newcommand{\R}{\ensuremath{\mathbb{R}}}
\newcommand{\Q}{\ensuremath{\mathbb{Q}}}
\newcommand{\N}{\ensuremath{\mathbb{N}}}
\newcommand{\Z}{\ensuremath{\mathbb{Z}}}

\newcommand{\file}[1]{\texttt{#1}}

\newcommand{\OneMax}{\textsc{OneMax}}

\usepackage{framed,mdframed}
\usepackage{multirow}
\usepackage{array}
\usepackage{bigdelim}
\usepackage{dsfont}
\usepackage{hyperref}

\definecolor{background}{RGB}{44,44,44}
\definecolor{string}{RGB}{230, 219, 116}
\definecolor{comment}{RGB}{117, 113, 94}
\definecolor{normal}{RGB}{248, 248, 242}
\definecolor{identifier}{RGB}{166, 226, 46}

\begin{document}

\newcommand{\note}[1]{{\color{red}\it #1}}

\title{IOHprofiler: A Benchmarking and Profiling Tool for Iterative Optimization Heuristics}
	\author{Carola Doerr$^1$, Hao Wang$^2$, Furong Ye$^2$, Sander van Rijn$^2$, Thomas B\"ack$^2$}
		\date{$^1$Sorbonne Universit\'e, CNRS, Laboratoire d'informatique de Paris 6 (LIP6), Paris, France\\
		$^2$ LIACS, Leiden University, Niels Bohrweg 1, 2333CA Leiden, The Netherlands\\ \vspace{2ex}
		\today}
\maketitle

\sloppy{
\begin{abstract}
IOHprofiler is a new tool for analyzing and comparing iterative optimization heuristics. Given as input algorithms and problems written in C or Python, it provides as output a statistical evaluation of the algorithms' performance by means of the distribution on the fixed-target running time and the fixed-budget function values. In addition, IOHprofiler also allows to track the evolution of algorithm parameters, making our tool particularly useful for the analysis, comparison, and design of (self-)adaptive algorithms.  

IOHprofiler is a ready-to-use software. It consists of two parts: an experimental part, which generates the running time data, and a post-processing part, which produces the summarizing comparisons and statistical evaluations. The experimental part is build on the COCO software, which has been adjusted to cope with optimization problems that are formulated as functions $f:\mathcal{S}^n \to \R$ with $\mathcal{S}$ being a discrete alphabet of integers. The post-processing part is our own work. It can be used as a stand-alone tool for the evaluation of running time data of arbitrary benchmark problems. It accepts as input files not only the output files of IOHprofiler, but also original COCO data files. The post-processing tool is designed for an interactive evaluation, allowing the user to chose the ranges and the precision of the displayed data according to his/her needs.   

IOHprofiler is available on GitHub at \url{https://github.com/IOHprofiler}.
\end{abstract}

\vspace{2ex}

\pagebreak
\tableofcontents
\pagebreak

\begin{figure*}
\centering
\fbox{\includegraphics[width=\linewidth]{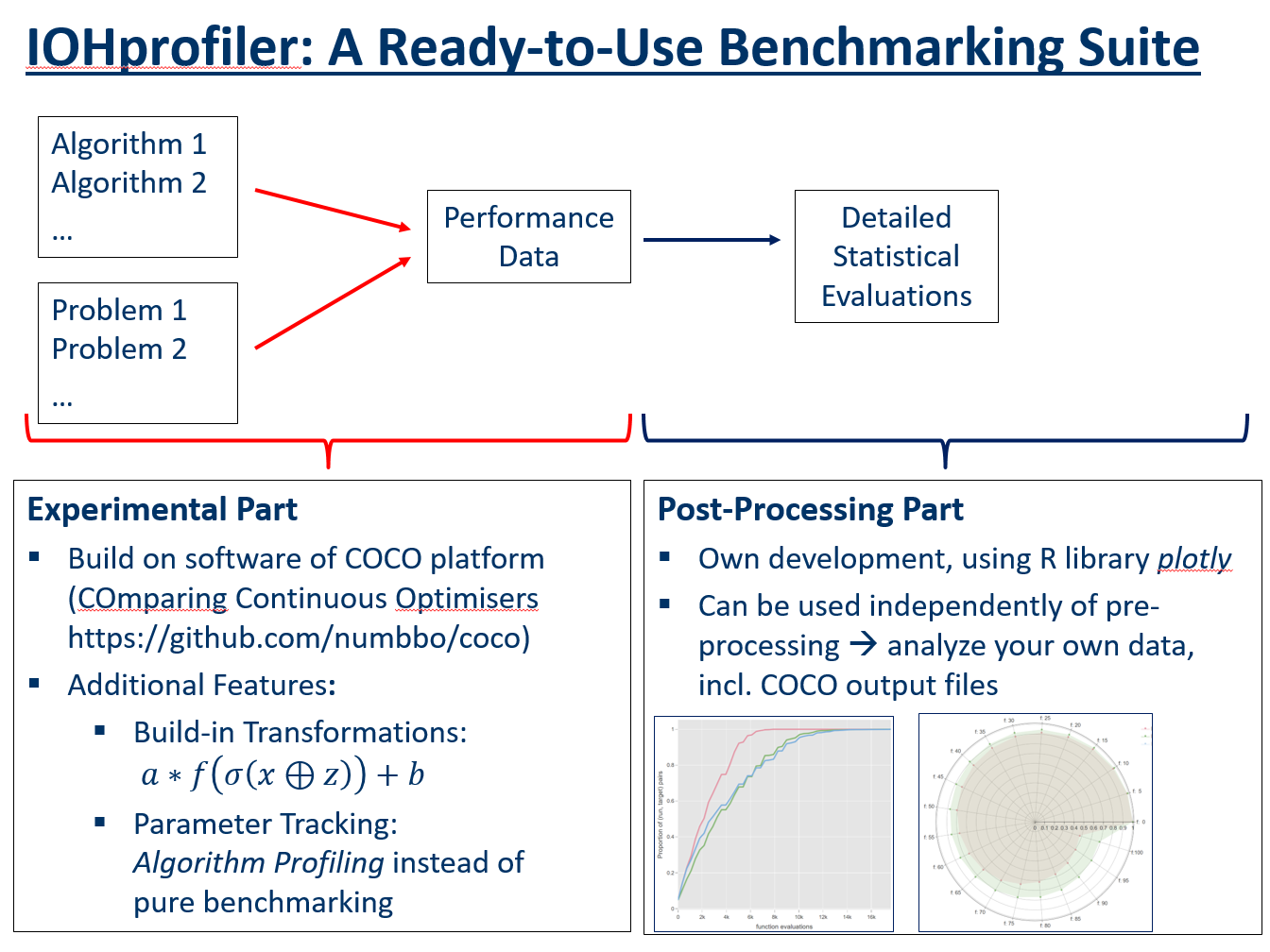}}
\caption{General Layout of \tool. The post-processing part can be used independently of the experimental part and supports the analysis of COCO output files.}
\end{figure*}
\vfill
\clearpage

\section{Introduction}\label{sec:intro}

The ultimate goal of research on optimization problems is the design of efficient problem solvers, which determine high-quality solutions at low cost. Thousands of new algorithms are suggested every year, and the questions of how their performances compare across different optimization problems, and of how far the underlying design ideas can be used to solve different types of optimization problems impose themselves. \emph{Benchmarking} addresses these questions in a principled way, by providing an empirical performance evaluation across different types of optimization problems. The task of designing suitable benchmarking environments is highly non-trivial, and comprises the following questions: 
\begin{enumerate}
	\item[Q1] Which type of optimization algorithms shall be compared? 
	\item[Q2] Which benchmark problems are most suitable for the comparison? 
	\item[Q3] Which performance measures should be used?
	\item[Q4] Apart from performance, which additional properties of the algorithms should be compared?
\end{enumerate}
In addition to these questions, a number of technical difficulties, such as the various programming languages in which the algorithms and problems are written or the platform on which the benchmark suite is executed also need to be addressed. 

The answer to any of the questions Q1-Q4 is quite subjective, and we cannot expect to reach consensus among the scholars and users of different optimization methods. When restricting to certain classes of optimization algorithms, however, it is possible to distill a number of common design principles. In the following, we briefly explain the choices and assumptions made by \tool. 

\subsection{Iterative Optimization Heuristics}
With respect to Q1, we focus in this work on \emph{iterative optimization heuristics} (IOH). As IOH we classify all algorithms which aim to find optimal solutions by an iterative search. That is, to optimize a problem $f:\mathcal{S} \to \R$, these algorithms proceed in rounds. In each round, the objective values of one or more solution candidates (\emph{search points}) $s^1, \ldots, s^{\lambda} \in \mathcal{S}$ are evaluated. Their function values $f(s^1), \ldots, f(s^{\lambda})$ are used to update the strategy by which the search points for the next round are generated. The search continues until a stopping criterion has been met, e.g., when a solution of a desired quality has been found, a time budget has been reached, or no significant progress could be observed in the last iterations.

The class of IOH subsumes \emph{local} search variants (including first/steepest ascent, variable neighborhood search, Simulated Annealing, Metropolis algorithms, etc.) and \emph{global} search heuristics such as evolutionary algorithms, (Quasi-)Monte Carlo algorithms, swarm intelligence, differential evolution, estimation of distribution algorithms, efficient global optimization, Bayesian optimization, etc. 

IOH are particular useful for the optimization of complex, high-dimensional, and large-scale optimization problems. They are---in par with mathematical programming---among the most frequently applied optimization routines in industrial and academic optimization.

\subsection{Real-Valued Optimization}
Providing an answer to Q2 is arguably the most subjective part of the decision process. We have chosen to take this question aside and to present a very general benchmarking environment which allows an in-depth comparison of IOH for arbitrary real-valued optimization problems.\footnote{At the moment, we assume only that the problems are static (i.e., $f$ does not change while being optimized) and noise-free (i.e., $f$ is a deterministic function). An extension to dynamic, noisy, and multi-objective optimization problems is under consideration. Users interested in such cases are invited to contact the authors to discuss how to modify \tool to cover such optimization problems.} That is, we do not intend in this work to discuss which problems are particularly suitable for the comparison of different IOH. Rather do we offer a tool that can be used to compare performance across functions of the user's choice.  

Since the algorithms need to know which search space they should operate on, we focus in the experimental part on problems defined over a discrete alphabet; i.e., we allow functions of the type $f:\mathcal{S}^n \to \R$ with $\mathcal{S} \subset \Z$ being a discrete alphabet of integers. Note that this class comprises in particular the (very broad) class of \emph{pseudo-Boolean optimization problems}, i.e., functions of the type $f:\{0,1\}^n \to \R$. 

The post-processing part does not make any assumptions on the type of problem. It can also be used to compare performance across arbitrary optimization problems $f:\R^n \to \R$. It notably accepts in particular data files produced by the original COCO software. 

The experimental part of \tool assume \emph{maximization} as objective. The post-processing part automatically detects from the best-so-far values whether minimization or maximization was the objective of the corresponding experiment.    

\subsection{Performance Measures}
With respect to Q3 we make the important assumption that the running time of the algorithms is dominated to a significant extend by the evaluation of solution candidates, so that all performance measures are based on the \emph{number of function evaluations.} Originally inspired by so-called \emph{black-box optimization} (which, in intuitive terms, assumes that the objective function is not accessed by the algorithm other than through the evaluation of solution candidates), these measures are today established performance indicators also in situations where algorithms \emph{do} have access to (and do make use of) instance data. In several streams of Computer Science the evaluation of a function value is considered a \emph{query} (with the idea that the objective value is queried from an oracle), and the number of function evaluations referred to as \emph{query complexity}. The advantage of evaluation-based performance measures is that they are independent of the machine on which they are executed. However, the user should keep in mind that query complexity can only give an accurate picture of CPU time when the latter is indeed determined to a large extend by the number of evaluated samples. 

As standard performance measures \tool provides information about the distribution of \emph{fixed-target} running times and \emph{fixed-budget} function values. These results include average values and quantiles, but also empirical cumulative distribution function (ECDF) curves, histograms, and empirical probability mass functions. All results are presented in an interactive format that allows the user to specify the granularity, the ranges, and the precision at which the results are displayed. All plots can be stored as png files, the data tables as csv files.

\subsection{Tracking Additional Information}
Addressing Q4, \tool can also be used to analyze the evolution of various algorithm parameters. The parameters to be tracked are specified by the user. For any of these parameters the user can obtain the same type of statistics as for the running time data. That is, the standard output of \tool includes in particular statistics (average, median, quantiles,...) about the parameter value at a given point in time (fixed-budget perspective) and at a given function value (fixed-target perspective). 

This profiling aspect can be of independent interest, as such information can be very useful for the design of suitable optimization heuristics. 

\subsection{GitHub page of IOHprofiler}
\tool can be downloaded from the GitHub page \url{https://github.com/IOHprofiler}. 

\subsection{Mailing List}
Users interested in receiving important updates about \tool can subscribe to a mailing list at \url{http://eepurl.com/dBahWb}. 

\subsection{User Support}
The development team of \tool can be reached by e-mail at \href{mailto:Carola.Doerr@mpi-inf.mpg.de}{Carola.Doerr@mpi-inf.mpg.de}\footnote{This address will be updated with a more generic one in the next version, but for now, please use this address.} 

Users can use this contact address to ask for support with the setup of \tool, but also to suggest new functionalities, different evaluation statistics, etc., or to provide feedback. 

\subsection{License and Main References}

The experimental part of \tool is build on the COCO software~\cite{hansen2016cocoplat}, available at \url{https://github.com/numbbo/coco}, and includes various modifications to adjust this tool to discrete optimization, to allow for the transformations described in Section~\ref{sec:functions}, to choose the granularity by which the data is stored, and to track algorithm parameters.  

The post-processing part is original work; the visualization of the results uses the R library \emph{plotly} from \url{https://plot.ly/}. 

\tool is governed by the BSD 3-Clause license.

\subsection{Structure of the Documentation}

In the following sections, we summarize the algorithm and problem requirements of \tool (Section~\ref{sec:requirements}), discuss the various options to set the precision of the performance evaluation (Section~\ref{sec:intro:precision}), provide an overview over the standard outputs generated by \tool (Section~\ref{sec:output}), and conclude with a number of extensions currently in preparation and planned for future releases of \tool (Section~\ref{sec:conclusions}). 

A step-by-step manual for the two experimental part of \tool is provided in Section~\ref{sec:pre} of the appendix. The manual for installing and running the post-processing part of \tool can be found on the aforementioned GitHub page \url{https://github.com/IOHprofiler}.

\section{Summary of Algorithm and Problem Requirements}\label{sec:requirements}
In this section we summarize the class of algorithms for which \tool can compare performance and discuss the type of optimization problems which are admissible.  

\subsection{Iterative Optimization Heuristics}
\label{sec:IOH}

The focus of \tool is on the performance analysis of \emph{iterative optimization heuristics} (IOH), the class of all algorithms that follow the structure of Algorithm~\ref{alg:IOH}. As mentioned in the introduction, this class comprises all sorts of randomized search heuristics, ranging from simple local hill-climbers to complex global search heuristics. The only important feature is that these heuristics do (not only) directly manipulate the problem data, but rather work in a trial-and-error fashion, in which the the function evaluation of search points $s \in \mathcal{S}$ is an integral part of the optimization routine.  

Note that in Algorithm~\ref{alg:IOH} all randomized decisions can be replaced by deterministic ones, so that the class of IOH also subsumes deterministic optimizers. 

\begin{algorithm2e}[t]%
	\textbf{Initialization:}\\
	\Indp
	$t \assign 0${\footnotesize{//iteration counter}}\;
	$\mathcal{H}(1) = \emptyset${\footnotesize{//search history}}\;
	\Indm
	\textbf{Optimization:} 
	\While{termination criterion not met}{
		$t \assign t+1$\;
		Based upon the search history $\mathcal{H}(t)$, choose a probability distribution on $\N$ and sample from it $\lambda(t)$; {\footnotesize{//number of samples to be queried in $t$-th iteration}}\\
		Based upon $\mathcal{H}(t)$ and $\lambda(t)$, choose a probability distribution $D(t)$ on $\mathcal{S}^{\lambda(t)}$; {\footnotesize{//strategy from which next solution candidates are generated}}\\ 
		From $D(t)$ sample $x^{(t,1)}, \ldots, x^{(t,\lambda(t))} \in \mathcal{S}$ and evaluate their function values $f(x^{(t,1)}),\ldots,f(x^{(t,\lambda(t))})$\;
		Build $\mathcal{H}(t+1)$ by selecting which of the samples $(x^{(t,1)},f(x^{(t,1)})), \ldots,(x^{(t,\lambda(t))},f(x^{(t,\lambda(t))}))$ and which of the samples from $\mathcal{H}(t)$ to keep in the search history\;
	}
\caption{Blueprint of an iterative optimization heuristic for the optimization of a function $f:\mathcal{S} \to \R$.}
\label{alg:IOH}
\end{algorithm2e}

\textbf{Counting function evaluations.} As mentioned above, an important assumption that we make about the algorithms is that their running time is determined (to a large extend) by the time needed to evaluate the solution candidates. We therefore regard in \tool only performance measures that are based on counting the number of function evaluations; either in a fixed-target or a fixed-budget sense. While the former answers the question how many evaluations are needed to identify a solution of a certain quality., the fixed-budget perspective addresses the complementary questions asking for the quality of the best solutions that can be identified within a given budget of function evaluations.   

\textbf{General-purpose vs. problem-aware algorithms.} Our original interest is in comparing \ioh that do not have any a priori knowledge about the (type of) optimization problem that they are facing. That is, we classically work in the aforementioned black-box setting, in which we assume that the algorithm only knows that the problem is a function $f:\mathcal{S}^n \to \R$; i.e., it ``knows'' in particular the domain (\emph{search space}) $\mathcal{S}^n$ and the co-domain $\R$ (and possibly some bounds on the co-domain). In the classic black-box optimization scenario, the only way to acquire knowledge about the problem instance $f$ is through the evaluation of potential solutions $s \in \mathcal{S}$. However, despite this initial motivation, \tool is nevertheless also suitable for the comparison and profiling of \emph{problem-aware heuristics,} which have been designed for a particular type of optimization problem. 

As we shall explain in the next subsection, \tool offers to test several \emph{problem instances} that are obtained from a given base problem through transformations of the search points and/or the function values. This allows the user to analyze, for example, if an algorithm is invariant with respect to problem representation and with respect to absolute function values.

\subsection{Admissible Benchmark Problems}\label{sec:functions}

As mentioned above, the experimental part of \tool assume maximization as objective, and allows arbitrary functions $f:\mathcal{S}^n \to \R$ with $\mathcal{S}$ being a discrete alphabet of integers. In contrast, the post-processing part of \tool does not make any further assumption about the type of problems for which the statistics are generated; all real-valued problems are admissible, and the objective can be either minimization or maximmization. 

Both parts assume that the optimization problem is \emph{static,} i.e., it does not change over time. We also assume that the evaluations are \emph{noise-free.} 

\textbf{Problem instances.} Instead of testing one particular problem $f$ only, the user can choose to run experiments on several problem instances that are obtained from $f$ through a set of transformations. In its most general form, \tool currently offers to return to the algorithm the values $af(\sigma(x\oplus z))+b$, where 
\begin{itemize}
	\item $a$ is a \textbf{multiplicative shift} of the function value,   
	\item $b$ is a \textbf{additive shift} of the function value,  
	\item $\oplus z: \{0,1\}^n \rightarrow \{0,1\}^n, (x_1,\ldots,x_n) \mapsto (x_1+z_1 \mod 2, \ldots, x_n+z_n \mod 2)$ is an \textbf{XOR-shift} of the search point,
	\item $\sigma: \mathcal{S}^n \rightarrow \mathcal{S}^n, (x_1,\ldots,x_n) \mapsto (x_{\sigma(1)}, \ldots, x_{\sigma(n)})$ is a \textbf{permutation} of the search point. Note here that, in abuse of notation, we identify the permutation $\sigma:[1..n] \to [1..n]$ with the here-defined re-ordering of the bit string.
\end{itemize}
Note that the ``$\oplus z$'' transformation is defined only for pseudo-Boolean problems $f:\{0,1\}^n \to \R$, but it can easily be extended to search spaces of the form $\{k_1,\ldots,k_1+r_1\} \times \ldots \times \{k_n,\ldots,k_n+r_n\}$.\footnote{Users interested in such an extension are invited to contact the authors to discuss a possible integration of such a transformation to \tool.}

The transformations defined above can be used to test if an algorithm behaves invariant under the proposed modifications. This is an often desired feature of \ioh. 

The user can choose if and which of the transformations are applied to his/her problem, cf. Section~\ref{sec:instances} for details. If no transformation is selected, the original $f(x)$-values are returned to the algorithm. 

We mention here already that, for convenience of the data analysis, the output files store the following four values: 
 \begin{itemize}
	 \item $f(\sigma(x\oplus z))$, the non-shifted function value of the search point evaluated in the corresponding iteration,
	 \item the best-so-far $f(\sigma(x\oplus z))$ value, 
	 \item $af(\sigma(x\oplus z))+b$, the shifted function value of the current search point; this is the value that the algorithm has access to, and
	 \item the best-so-far $af(\sigma(x\oplus z))+b$ value.
 \end{itemize}

\section{Precision of the Analysis and Data Format}
\label{sec:intro:precision}

A sound statistical comparison of algorithms requires a substantial amount of data. For the standard output of \tool, we track for selected evaluations the number of search points evaluated up to this iteration, the (transformed and the original) function value of the solution under evaluation, and the (transformed and original) function value of the best-so-far solution. Apart from this information, \tool can store additional data, such as the parameters that determine the exact structure of the algorithm. Typical examples for such parameters are the radius at which new solution candidates are sampled (e.g., the \emph{mutation rate}), the number of offspring evaluated in the present iteration, and parameters that determine the selection of the points to keep in the memory. The user selects which algorithm parameters are tracked, cf. Section~\ref{sec:pre} for details. 

An example for a standard output file be seen in Figure~\ref{fig:outputfile}. 
\begin{figure}[t]
\centering
\includegraphics[width=\linewidth]{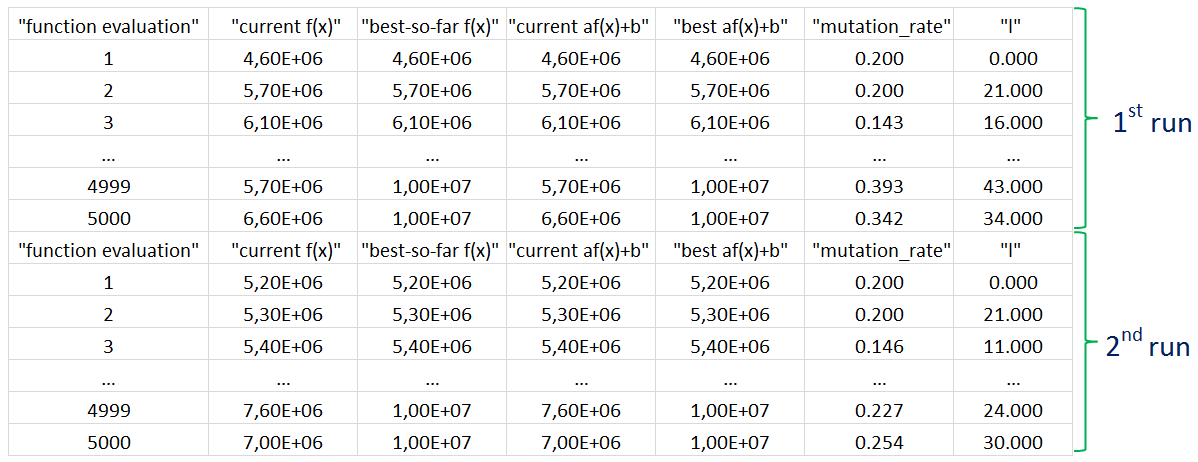}
\caption{The output of the experimental part of \tool are log-files with information about the current and best-so-far function values and possibly additional algorithm parameters, the mutation rate and parameter ``l'' in this case. These files are the input for the post-processing part of \tool.}
\label{fig:outputfile}
\end{figure}

The interval at which data is stored is chosen by the user. \tool allows for the following options. 
A detailed description how to select this granularity is provided in Section~\ref{sec:pre}.
\begin{itemize}
	\item \textbf{Complete tracking (*.cdat files):} This data file provides the highest granularity, by storing the above-described information for each function evaluation. 
	\item \textbf{Interval tracking (*.idat files):} The user specifies a step size $\tau$. Data is stored for every $\tau$-th function evaluation.  
	\item \textbf{Target-based tracking (*.dat files):} These data files store data for each iteration in which the best-so-far function value improves. 
	\item \textbf{Time-based tracking (*.tdat files):} In this data file, records are written when the user-specified running time budgets are reached. These running time budgets are evenly spaced in the log-$10$ scale, taking the form ${v10^i \mid i = 0, 1,2, \ldots}$ or ${10^{i / t} \mid i=0, 1,2,\ldots}$. Here, $v$ and $t$ can be set by the user.
\end{itemize}
With the current experimental setup, the *.dat data format is always generated, the other three are optional. 

The structure of the output files follows very closely that of the COCO environment: For each tested algorithm, a separate folder \emph{Algorithm1.zip} is created; the name of this folder can be chosen by the user, see Section~\ref{sec:pre} for details. 
In this folder we find for each tested benchmark function a ``.info'' file, e.g., 
\texttt{IOHprofiler\_f2\_i1.info} (where the part ``\_f2'' indicates the tested function and the part ``\_i1'' the smallest index of the tested instances of this benchmark problem). 
\vspace{2ex} 
\includegraphics[width=\linewidth]{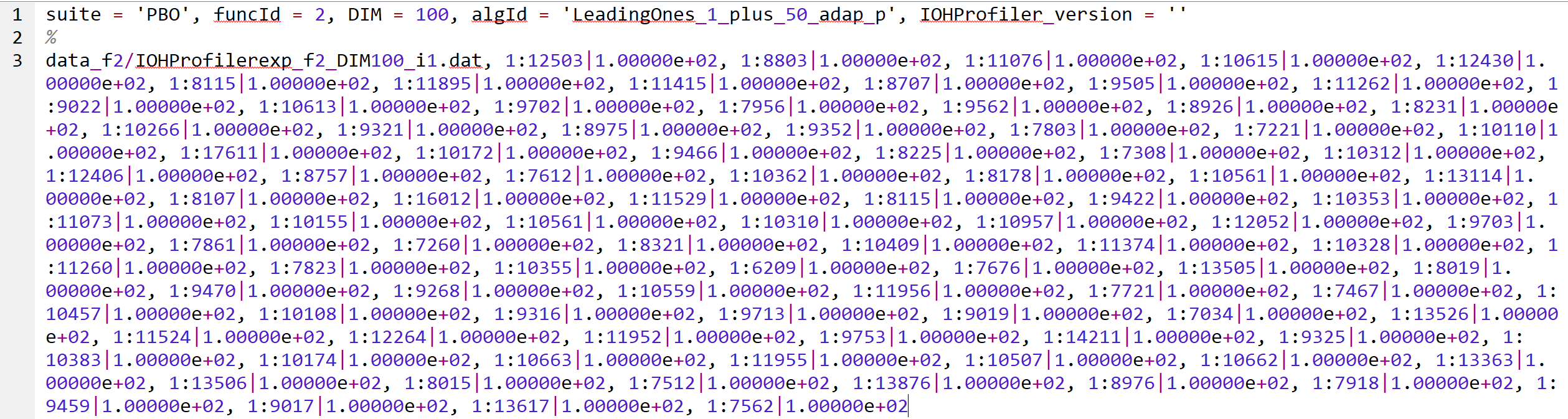}
\vspace{2ex}
This file contains the following information: 
\begin{itemize}
	\item in the first line we store the name of the benchmark suite (a \emph{suite} is a collection of benchmark functions), the ID of the benchmark function, the dimension for which experiments have been conducted, the name of the algorithm, and information about the version of the IOHprofiler.
	\item the second line is an empty line containing only the symbol \%. The user can use this line to record some information about the algorithm or the experiment. This information can be specified in the configuration file.
	\item the subsequent lines specify the path where the actual runtime data is located (in the example of the screenshot above, this is the file \texttt{IOHprofilerexp\_f2\_DIM100\_i1.dat} in folder \emph{data\_f2}. Thereafter, it is recorded for each run (100 in the example) how many lines of data points have been stored, along with the final best-so-far value of the respective run. In the example above, 12\,503 data points have been stored for the first run, and the best found solution had a function value of $100$.  
\end{itemize}
When several dimensions have been tested, the corresponding information above is written into the \texttt{IOHprofiler\_f2\_i1.info} one below the other. 

The performance data is stored in sub-folders; one subfolder for each tested benchmark function. 
In these sub-folders the different data files specified above can be found, generic names for these files are \texttt{IOHprofiler\_f2\_DIM1000\_i1.dat} for a *.dat file containing performance data from an experiment on the $1\,000$-dimensional function f2. Results for different dimensions are stored in the same folder.

\section{Supported Performance Analyses}
\label{sec:output}

We recall that the objectives of \tool tool are two-fold. On the one hand, it aims to contribute to a statistically sound comparison of iterative heuristics for pseudo-Boolean optimization problems. This is the \textbf{benchmarking aspect} of \tool. An important motivation for algorithm benchmarking is the desire to generate insights that can be used for the design of efficient optimizers. To this end, it is not only important to understand well how the algorithms perform on different types of optimization problems; not less important is to analyze how the states of the algorithm itself evolve over time. To address this question, \tool allows to track the evolution of the key parameters that determine the algorithm. The evaluation of these parameters covers the \textbf{profiling aspect} of \tool.

We describe in this section the standard outputs that \tool generates. The results are grouped into three categories: 
\begin{enumerate}
	\item \textbf{Fixed-Target Results,} described in Section~\ref{sec:post:target}: This section provides summarizing statistics covering the fixed-target perspective of performance evaluation. That is, the results in this section mainly address the question how much ``time'' (i.e., function evaluations) is needed to obtain a solution of a desired target quality. 
	\item \textbf{Fixed-Budget Results,} Section~\ref{sec:post:budget}: Covering the fixed-budget perspective, these outputs present statistics for the quality of the search points obtained within a given budget of function evaluations. That is, the results in this section mainly address the question how good the search points are that a user can expect to see within a given time frame (where ``time'' refers again to the number of evaluations).
	\item \textbf{Algorithm Parameters,} Section~\ref{sec:post:params}: This section provides details about the evolution of the algorithm parameters that the user specified to be tracked during the experimental part.
\end{enumerate}
 
\textbf{Performance Measure in Preparation:} \tool currently does not perform statistical tests, nor comparing results over various problem dimensions, nor performance aggregation over several benchmark problems. These measures are currently in preparation, and will be made available shortly. Note, however, that in addition to the summarizing statistics detailed in the next subsections, \tool provides for each section the option to store sorted raw data, which may be convenient for computing additional performance measures, statistical tests, etc. Users interested in a discussion which additional performance measures to include as a standard output, are asked to get in touch with the \tool developers. 

\subsection{Notation and Basic Terminology}\label{sec:notation}

Before we present the various outputs, we briefly discuss the terminology used in the remainder of this section. We recall that we assume \emph{maximization} as objective. 

For every algorithm $A$ and every function $f$, we denote by 
\begin{itemize}
	\item $T(A,f,v,i)$ the number of function evaluations that have been performed in run $i$ until and including the first evaluation of a search point $x$ satisfying $f(x)\ge v$; i.e., the ``time'' needed by algorithm $A$ in run $i$ to reach for function $f$ a solution $x$ with \emph{target value} at least $v$.
	\item $V(A,f,t,i):=\max\{f(x^{(j)}) \mid j \in \{1,\ldots,t\}\}$, the function value of the best among the first $t$ evaluated solution candidates in run $i$. The variable $t$ is referred to as \emph{budget}. 
\end{itemize}

The values $T(A,f,v,i)$ and $V(A,f,t,i)$ are aggregated over the $r$ independent runs to \emph{fixed-target running times} and \emph{fixed-budget function values}, respectively. Note that $T(A,f,v,\cdot)$ and $V(A,f,t,\cdot)$ are random variables, and $T(A,f,v,i)$ and $V(A,f,t,i)$ samples thereof. 

Among the most classic performance measures are the \emph{mean} values $\mathbb{E}[V(A,f,t)]$ and $\mathbb{E}[T(A,f,v)]$ of the distributions $V(A,f,t,\cdot)$ and $T(A,f,v,\cdot)$. We approximate these expected values by the empirical averages over all $r$ independent runs, and abbreviate: 
\begin{itemize}
\item $\widehat{\mathbb{E}}[T(A,f,v)]:=\sum_{i=1}^{r}{T(A,f,v,i)/r}$, the average budget needed to find a solution of quality at least $v$.
	\item $\widehat{\mathbb{E}}[V(A,f,t)]:=\sum_{i=1}^{r}{V(A,f,t,i)/r}$, the average quality of the best solution found within a budget of $t$ function evaluations.
	\end{itemize}

When the variables $T(A,f,v,\cdot)$ and $V(A,f,t,,\cdot)$ are not concentrated and/or not symmetric, average values can be misleading. \tool therefore also computes different quantiles of these distributions. To this end, the values $V(A,f,t,1), \ldots, V(A,f,t,r)$ [and $T(A,f,v,1), \ldots, T(A,f,v,r)$, respectively] are sorted in non-decreasing order. We denote by $V(A,f,t,i:r)$ [and $T(A,f,v,i:r)$, resp.] the $i$-th element of the resulting sequence. For any $1 \le p \le 100$, the \emph{$p$-th percentile} of the distributions are estimated as
	$$\widehat{\Q}[T(A,f,v),p]:=T(A,f,v,\lfloor pr/100 \rfloor) \text{, and } \widehat{\Q}[V(A,f,t),p]:=V(A,f,t,\lfloor pr/100 \rfloor),$$
respectively.

In addition to these values, it is also interesting to accumulate the running time data into ECDF curves. ECDF stands for \emph{empirical cumulative distribution function.} Again we have to distinguish between the fixed-target and the fixed-budget perspective:
\begin{itemize}
	\item In the fixed-target perspective, an ECDF curve requires to select a set $\{v_1,\ldots,v_d\}$ of target values. The corresponding ECDF curve shows for each budget $t$ the fraction $\{(i,v_j) \mid 1 \le i \le r, 1 \le j \le d\}/(rd)$ of the (run, target value) pairs $(i,v_j)$ that satisfy that $V(A,f,t,i) \ge v_j$. That is, 
	$\widehat{F}(t) = \frac{1}{dr}\sum_{j=1}^{d}\sum_{i=1}^{r}\mathbf{1}_{V(A,f, t, i) \geq v_j},$
		where $\mathbf{1}_{\mathcal{C}}$ denotes the indicator variable, which is one when the condition $\mathcal{C}$ is satisfied.
	\item Likewise, in the fixed-budget perspective, the user selects a set $\{t_1,\ldots,t_d\}$ of budgets. The corresponding ECDF curve shows for each target value $v$ the fraction $\{(i,t_j) \mid 1 \le i \le r, 1 \le j \le d\}/(rd)$ of the (run, budget) pairs that satisfy that $T(A,f,v,i) \le t_j$.   
\end{itemize}
 
\subsection{Linking the Data Files}\label{sec:post:upload}

In the \textbf{upload} tab of the post-processing part, the user provides the links to the folders containing the performance data that shall be analyzed. Figure~\ref{fig:post:upload} shows this tab. The user can select whether his/her data is in the format of the \tool experimentation part or in the COCO format. Toggling the efficient mode results in a faster computation of the results, at the cost of precision. After choosing the data file to be uploaded to the tool, the \emph{Data Processing Promt} on the right records which data  has been identified; in the example 100 runs for the 100-dimensional version of function \emph{f2}. The list of processed data at the bottom summarizes this information in table format.  
\begin{figure}[t]
\centering
\includegraphics[width=\linewidth]{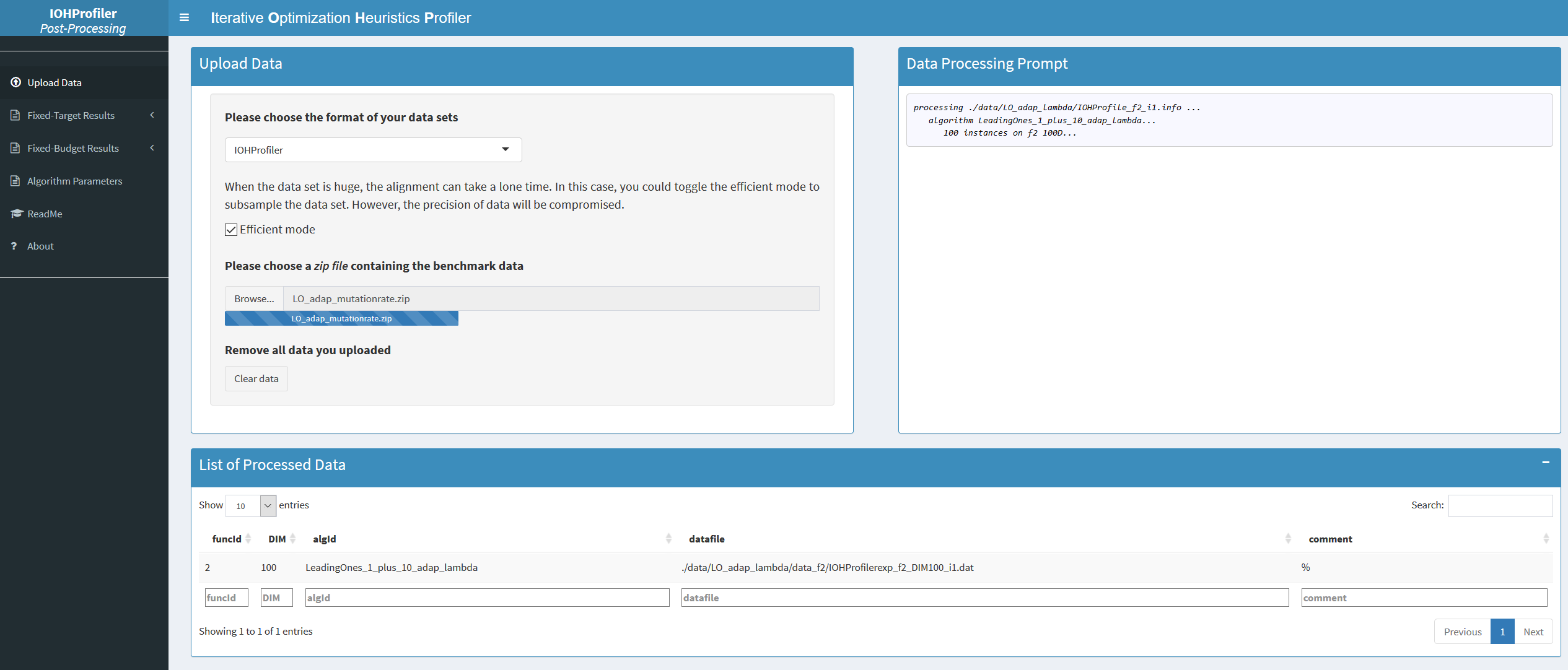}
\caption{In the \emph{upload} tab, the user provides the links the performance data.}
\label{fig:post:upload}
\end{figure}

\subsection{Fixed-Target Results}\label{sec:post:target}

The fixed-target section has four different subsections (``\emph{tabs}''): 
\begin{itemize}
	\item `Data Summary': this tab provides tables with the fixed-target running time statistics, as well as tables with the sorted raw values $T(A,f,v,i)$ of the individual runs. See Section~\ref{sec:target:data} for details.
	\item `Expected Runtime': an interactive plot illustrates the fixed-target running times. The user can choose to display mean and/or median values along with the standard deviations. The user also selects the algorithms which are displayed, the range for which the fixed-target statistics are computed, and whether or not the axes are  scaled logarithmically. Confer Section~\ref{sec:target:runtime} for details.
		\item `Probability Mass Function': interactive histograms show the distribution of the values $T(A,f,v,i)$ for target values $v$ selected by the user. Furthermore, an approximation for the empirical probability mass function is provided in this tab, cf. Section~\ref{sec:target:pmf}.
	\item `Cumulative Distribution': ECDF curves are computed for target values specified by the user. A spider-plot shows the area under the ECDF curves for different target values. In addition, ECDF curves for individual target values can be shown, cf. Section~\ref{sec:target:ECDF}.
\end{itemize}

\subsubsection{Fixed-Target: `Data Summary'}\label{sec:target:data}

Figure~\ref{fig:post:target:data} shows the upper part of the `Data Summary' tab. The user can set the range and the granularity of the results in the box on the left. The table shows fixed-target running times for evenly spaced target values. More precisely, for each (algorithm $A$, target value $v$) pair the table provides 
\begin{itemize}
	\item runs: the number of runs of algorithm $A$ in which at least one solution $x$ satisfying $f(x)>v$ has been found, 
	\item mean: $\widehat{\mathbb{E}}[T(A,f,v)]$, the average number of function evaluations needed to find a solution of function value at least $v$, 
	\item median, 2\%, 5\%, …: the quantiles $\widehat{\Q}[T(A,f,v),p]$ of these first-hitting times.
\end{itemize}

\begin{figure}[t]
\centering
\includegraphics[width=\linewidth]{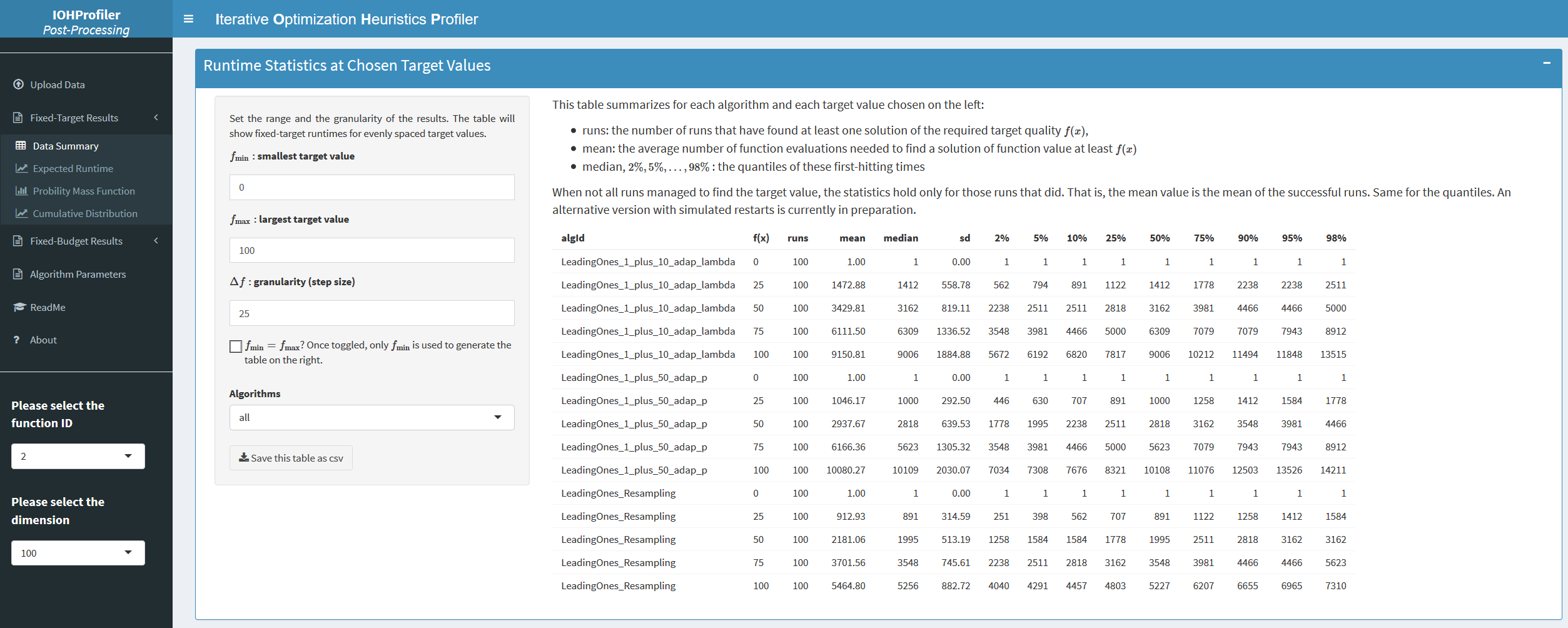}
\caption{`Fixed-Target Results: Data Summary': running time statistics at chosen target values.}
\label{fig:post:target:data}
\end{figure} 

The sorted raw data used to compute the summarizing statistics can be downloaded from the \texttt{Original Runtime Samples} section on the bottom of the `Data Summary' tab. For each target value $v$ selected in the options box on the left, the table shows, for each algorithm $A$ and each run $i$, the number $T(A,f,v,i)$ of evaluations performed by the algorithm until it evaluated for the first time a solution $x$ of quality at least $v$. The user can choose between a vertical and an horizontal alignment of the data; Figure~\ref{fig:post:target:samples} shows the wide variant. These tables can be stored as csv files.
\begin{figure}
\centering
\includegraphics[width=\linewidth]{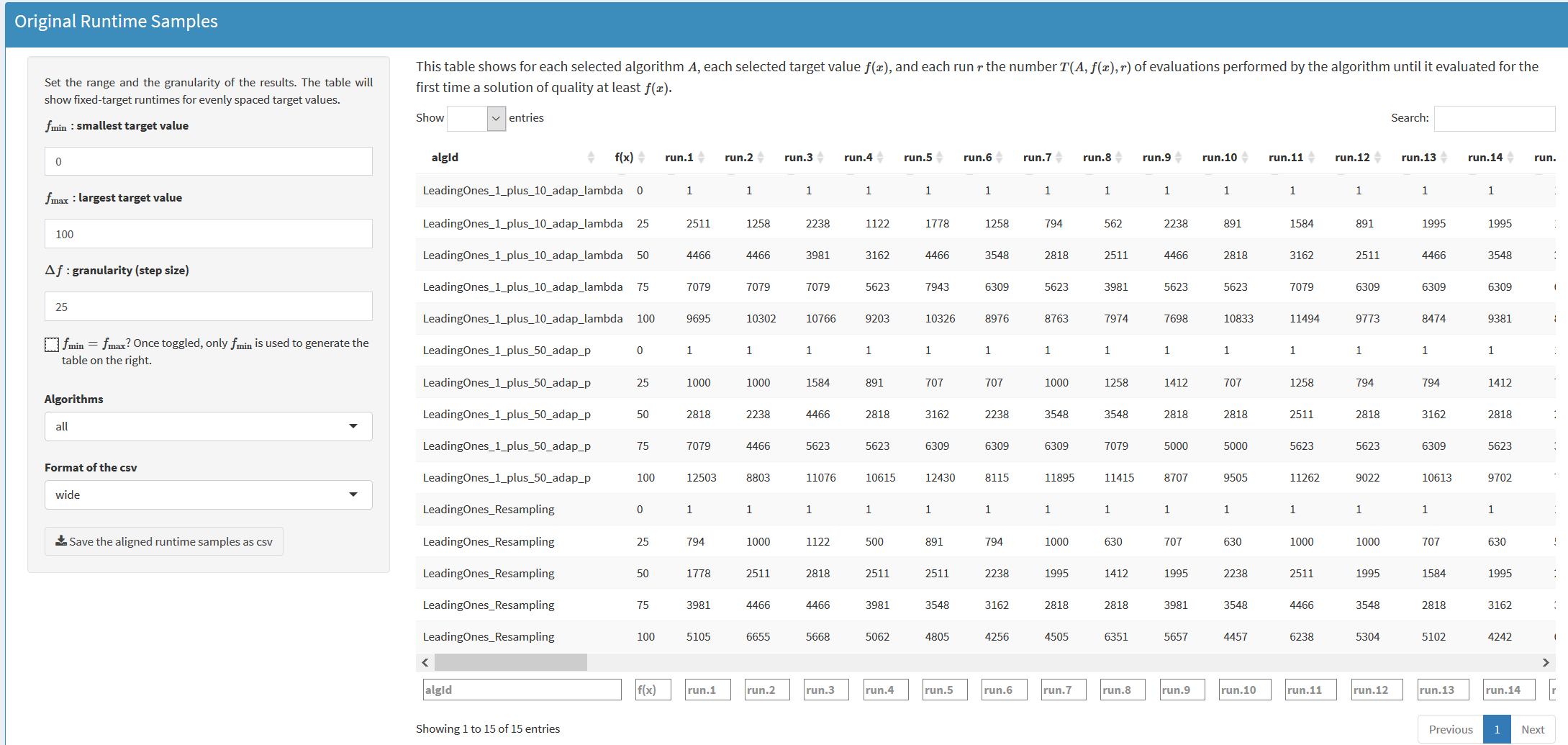}
\caption{`Fixed-Target Results: Data Summary': sorted first hitting times $T(A,f,v,i)$.}
\label{fig:post:target:samples}
\end{figure}

\subsubsection{Fixed-Target: `Expected Runtime'}\label{sec:target:runtime}

The average, median, and standard deviations of the running time samples are depicted against the best-so-far objective values. The displayed elements can be switched on and off by clicking on the legend on the right. This also allows the user to select the algorithms for which the results are shown. Some display options, including the option to store the picture as a png file, appear when moving the mouse over the picture. Detailed numbers appear when hovering the mouse over the curves, cf. Figure~\ref{fig:post:target:runtime}.

\begin{figure}[t]
\centering
\includegraphics[width=\linewidth]{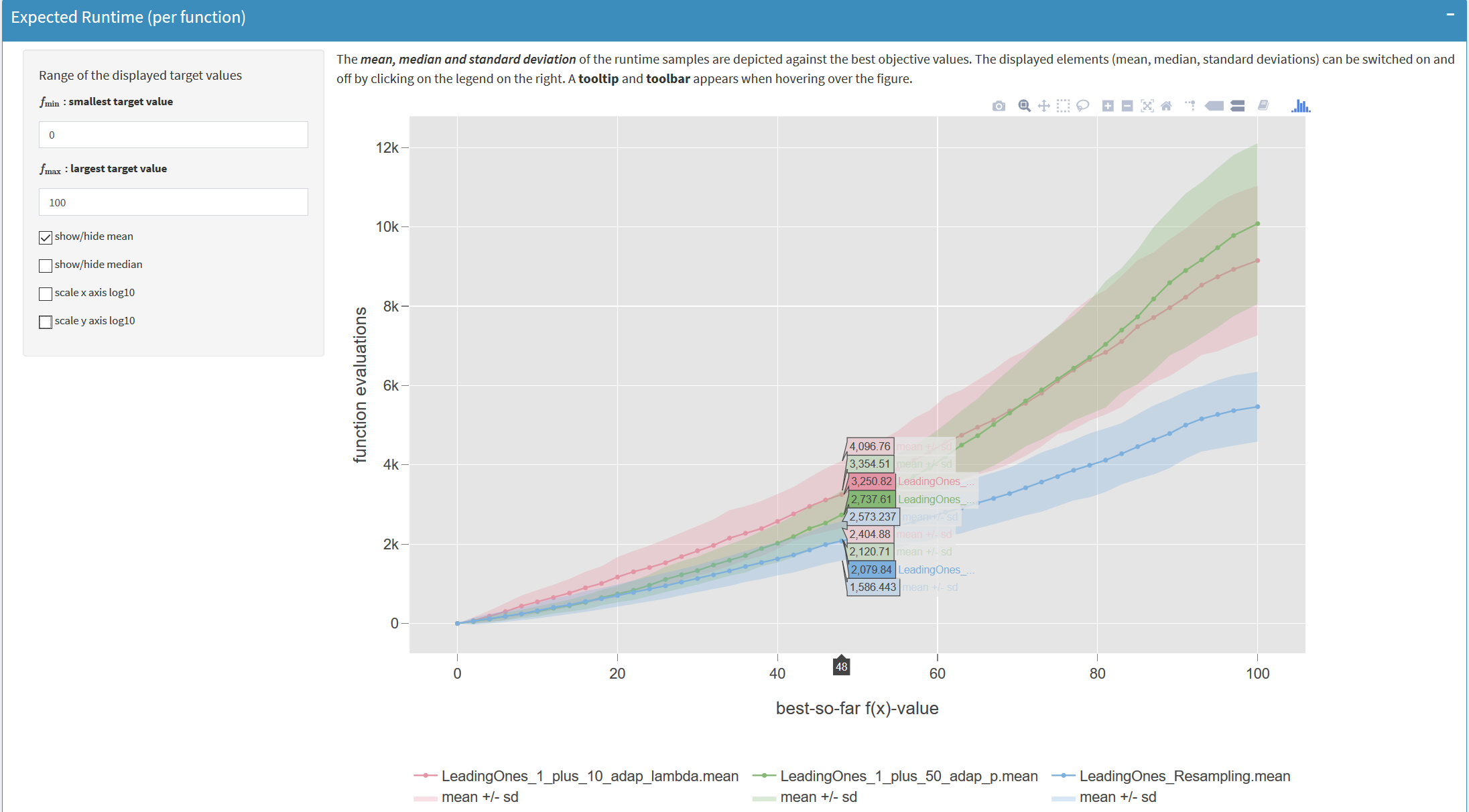}
\caption{`Fixed-Target Results: Expected Runtime': Average and median fixed-target running times.}
\label{fig:post:target:runtime}
\end{figure}

\subsubsection{Fixed-Target: `Probability Mass Function'}
\label{sec:target:pmf}

The third tab of the fixed-target section provides, for a target value selected by the user, histograms of the running time samples and an approximation of the probability mass function. 

For a selected target value $v$ the histogram, displayed in Figure~\ref{fig:post:target:histogram}, shows for each range $[t,t+1)$ the number of runs $i$ satisfying $t \le T(A,f,v,i) < t+1$. The bin sizes $[t,t+1)$ are chosen automatically according to the so-called Freedman–Diaconis rule, by which the bin size is set to $(\widehat{\Q}[T(A,f,v),75]-\widehat{\Q}[T(A,f,v),25])/\sqrt[3]{r}$. Note that the displayed algorithms can be selected again by clicking on the legend on the right. The user has two options: an overlayed display, where all algorithms are displayed in the same plot, or a separated one, in which each algorithm is displayed in an individual chart.  

\begin{figure}
\centering
\includegraphics[width=\linewidth]{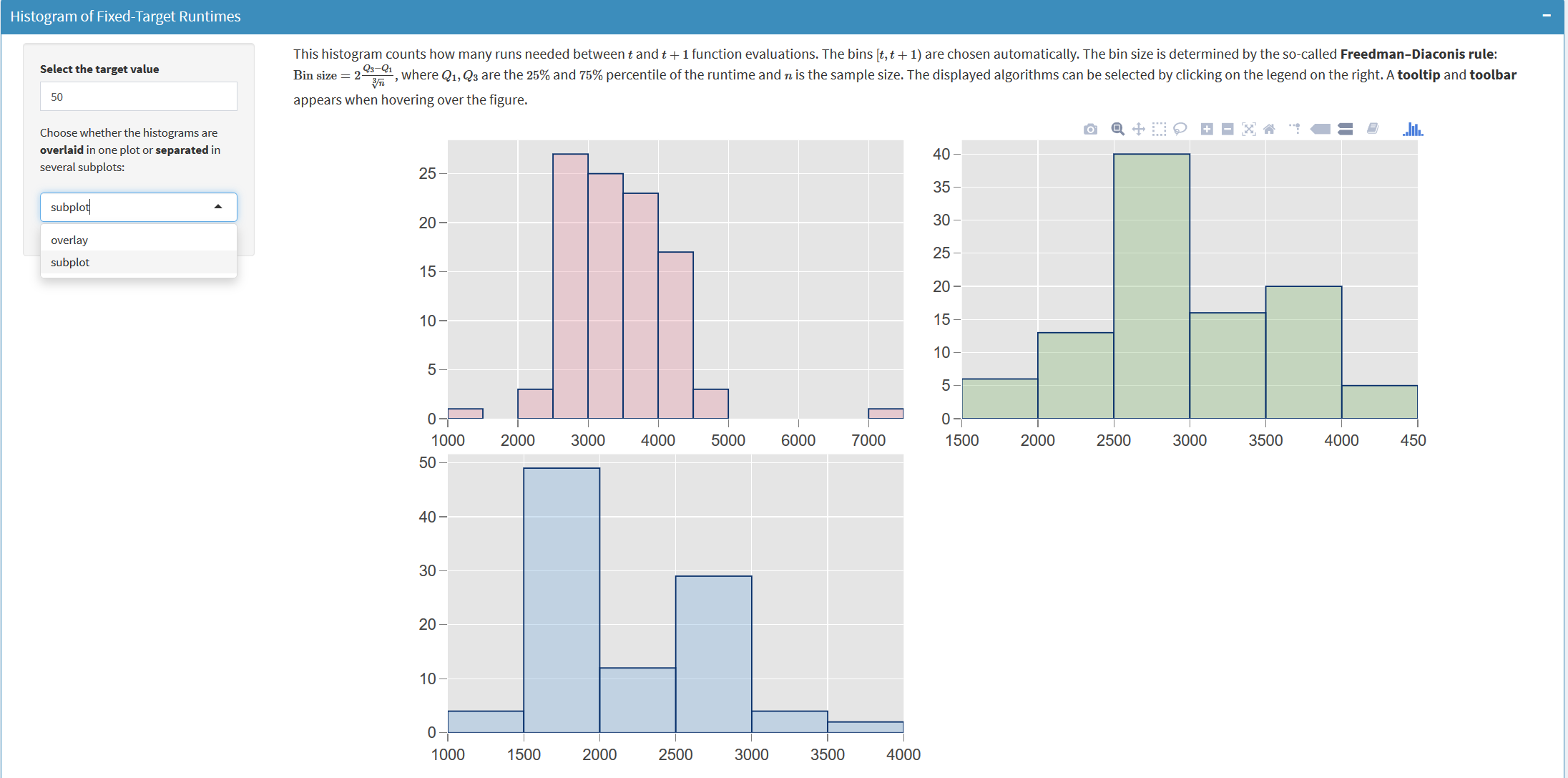}
\caption{`Fixed-Target Results: Probability Mass Function': Histograms for the fixed-target running times $T(A,f,v,i)$.}
\label{fig:post:target:histogram}
\end{figure}

Finally, the estimation of the probability mass function (cf. Figure~\ref{fig:post:target:pmf}) may be useful to get a better idea of how the values $T(A,f,v,i)$ are distributed for a given target value $v$. The user can opt to show all individual values $T(A,f,v,1), \ldots, T(A,f,v,r)$, or only the approximated probability mass function. Note, however, that the latter is just an approximation, which estimates the probability mass function by treating the running times as continuous variables.\footnote{Strictly speaking, this method gives imprecise estimations when there are many duplicated values. Improvements are planned for the future version.} Note also that in the example of Figure~\ref{fig:post:target:pmf} many data points seem aligned, this might be caused by turning (in the upload tab) the ``efficient mode'' on, in which the raw data set is trimmed.  

\begin{figure}
\centering
\includegraphics[width=\linewidth]{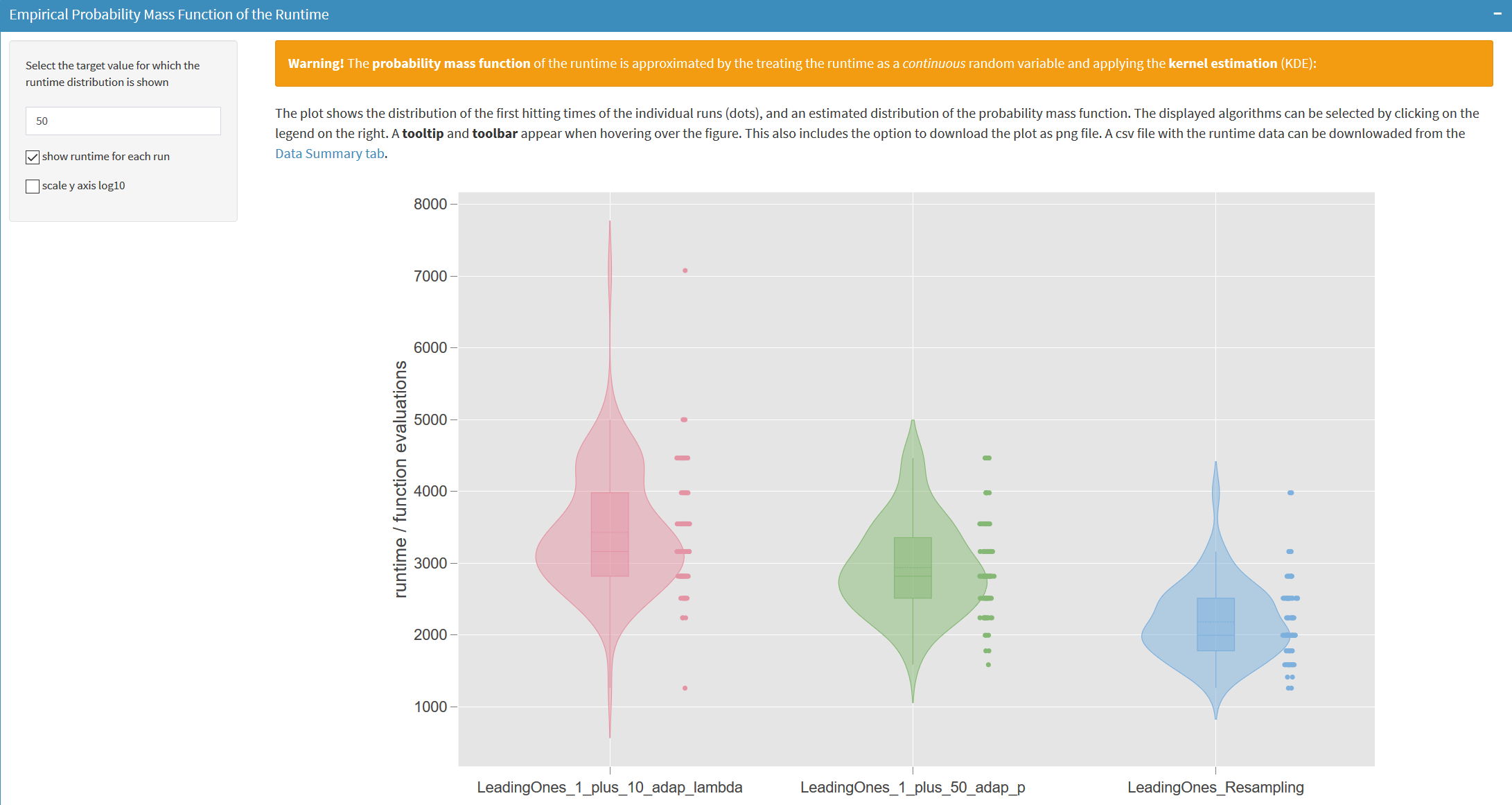}
\caption{`Fixed-Target Results: Probability Mass Function': approximated probability mass function for the fixed-target running times $T(A,f,v,i)$.}
\label{fig:post:target:pmf}
\end{figure}

\subsubsection{Fixed-Target: `Cumulative Distribution'}\label{sec:target:ECDF}

This tab provides ECDF curves and information about the area under the ECDF curves. For the aggregated ECDF curves, the user selects a range of the target values and the steps at which the data is displayed. Selecting as $f_{\min}=0$, $f_{\max}=100$, and $\Delta=10$ as in the example of Figure~\ref{fig:post:target:ECDF}, the ECDF curves for target values $0,10,20,30,...,100$ are computed. 
When $r$ independent runs have been performed, the ECDF curves thus show the fraction of all $11r$ (run, target value) pairs 
$\{(i,10j) \mid 1 \le i \le r, 0 \le j \le 10\}$ that satisfy for a given $t$ that $T(A,f,10j,i) \le t$. 
In the example of Figure~\ref{fig:post:target:ECDF}, for Algorithm \emph{LeadingOnes\_resampling} (blue curve) this is the case for around $75\%$ of the pairs after $t=4\,000$ function evaluations. For Algorithm \emph{LeadingOnes\_1\_plus\_50\_adap\_p} (green curve) the fraction is 58\%. 
 
\begin{figure}
\centering
\includegraphics[width=\linewidth]{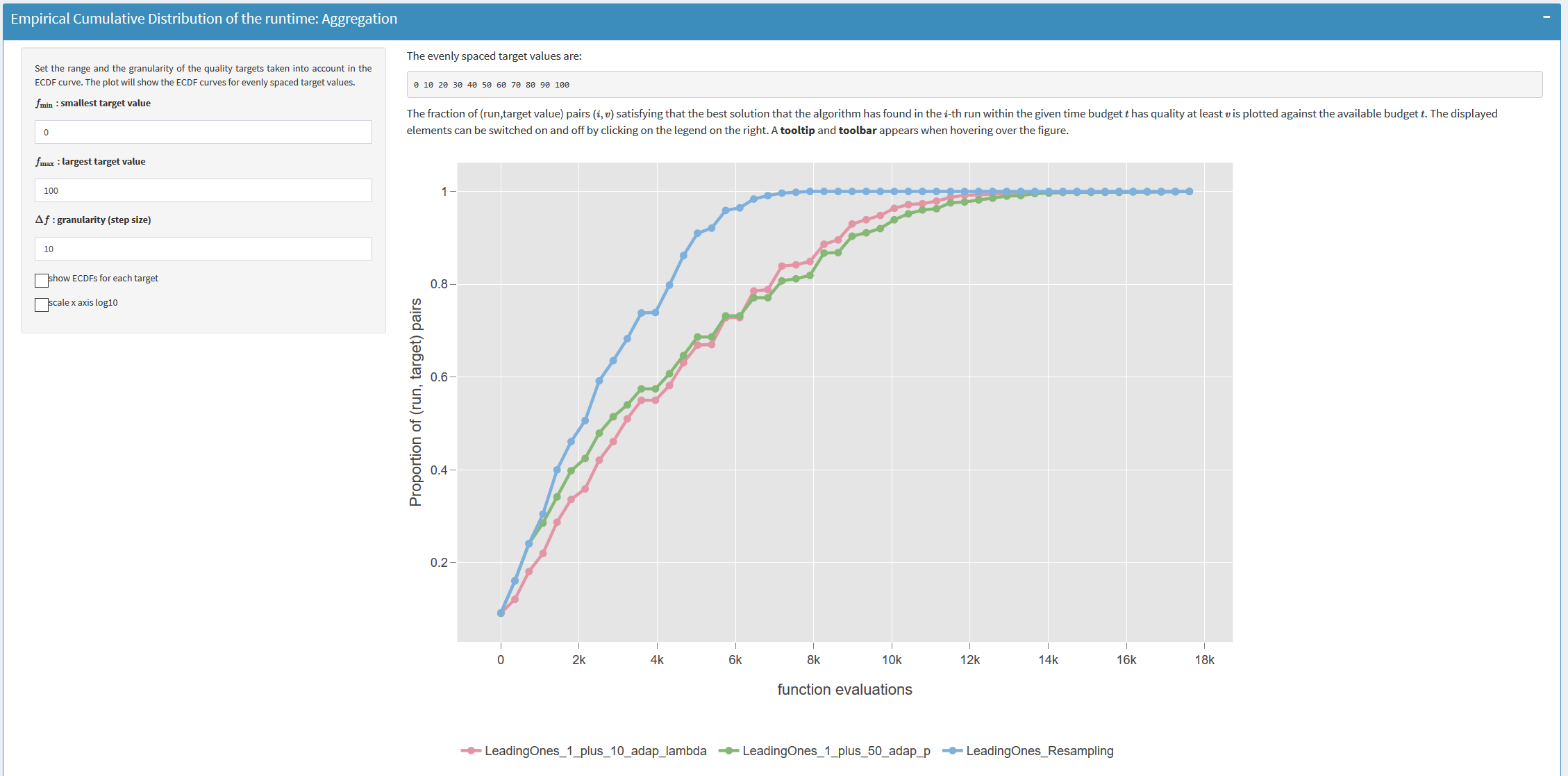}
\caption{`Fixed-Target Results: Cumulative Distribution': Aggregated ECDF curves for selected target values.}
\label{fig:post:target:ECDF}
\end{figure}

An ideal algorithm would sample the maximal function value $f_{\max}$ in the first step. This algorithm would have a 100\% score for all budgets $t$. In practice, such an algorithm does not exist, but it serves as a theoretical upper bound and we use the area under its curve to normalize the areas under the curves of the tested algorithms. The radar-like plot in the `Area under the ECDF' part of the `Cumulative Distribution' tab displays these normalized values for the (equally-spaced) target values chosen by the user, cf. Figure~\ref{fig:post:target:area}. 

\begin{figure}
\centering
\includegraphics[width=\linewidth]{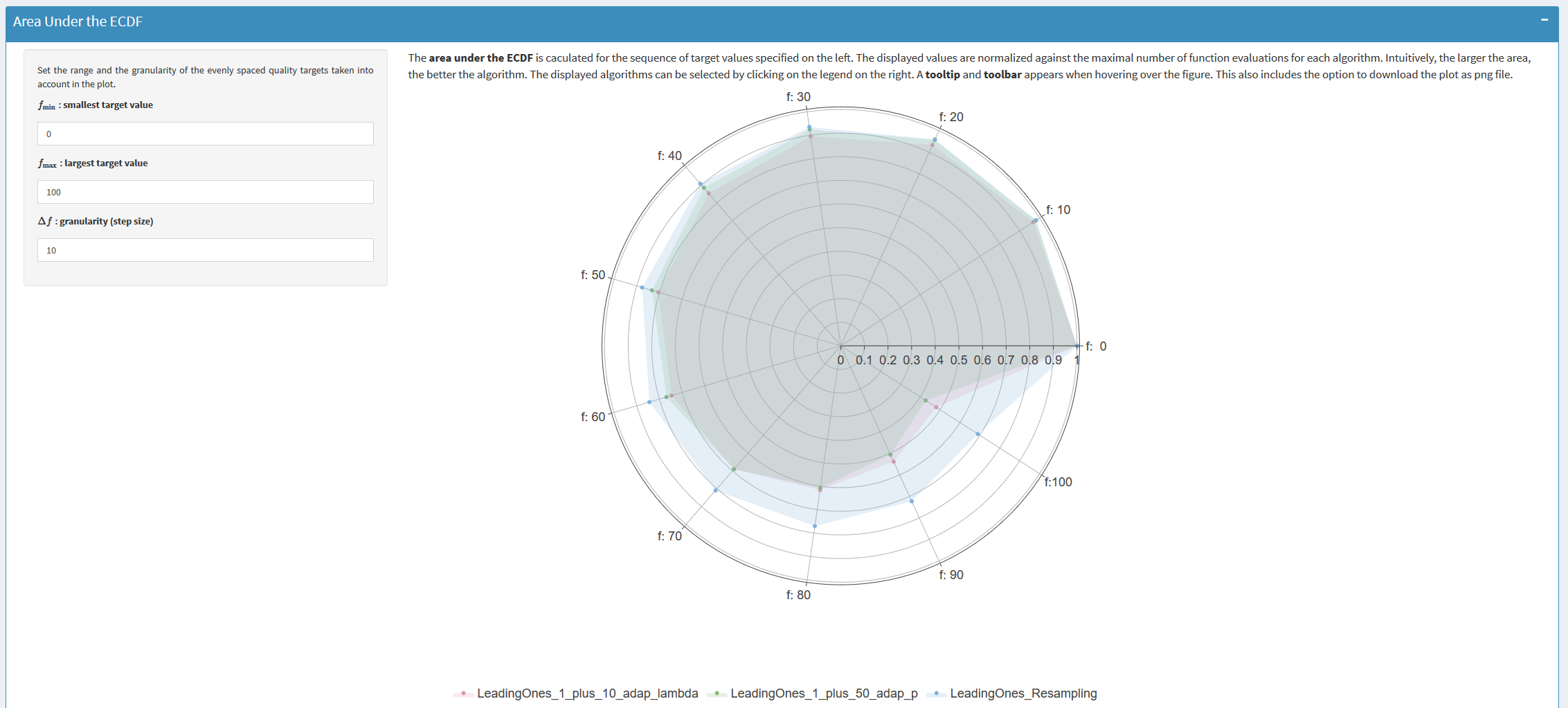}
\caption{`Fixed-Target Results: Cumulative Distribution': Area under the ECDF curves for selected target values, normalized by the area of the theoretically optimal algorithm evaluating a point with function value $\ge f_{\max}$ in the first iteration.}
\label{fig:post:target:area}
\end{figure}

ECDF curves for individual targets are available in the `Single Target' section of the `Cumulative Distribution' tab. An example is shown in Figure~\ref{fig:post:target:ECDFsingle}.

\begin{figure}
\centering
\includegraphics[width=\linewidth]{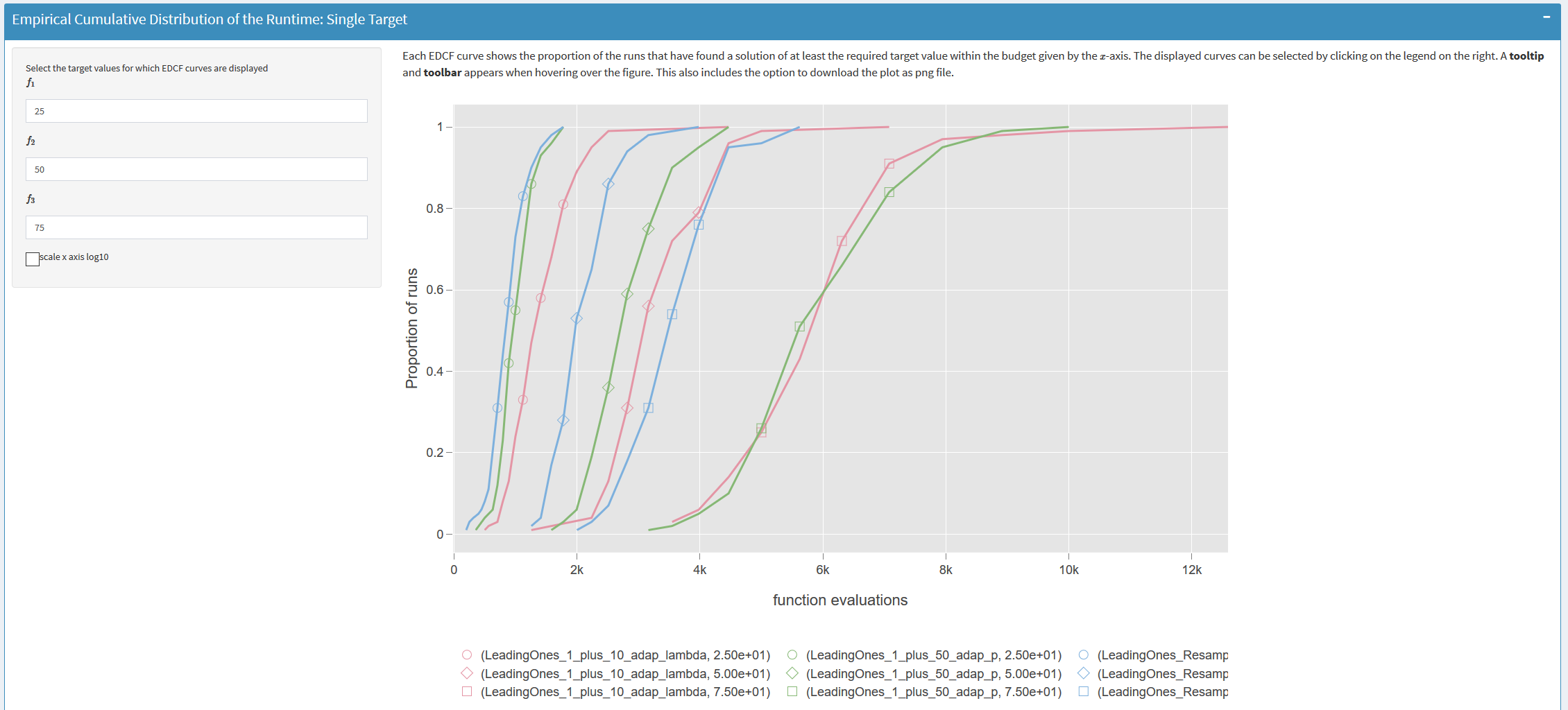}
\caption{`Fixed-Target Results: Cumulative Distribution': ECDF curves for individual target values.}
\label{fig:post:target:ECDFsingle}
\end{figure}

\subsection{Fixed-Budget Results}\label{sec:post:budget}

The fixed-budget section has the same four tabs as the fixed-target section:
\begin{itemize}
	\item `Data Summary': tables with fixed-budget running time statistics and sorted raw values $V(A,f,t,i)$ of the individual runs, 
	\item `Expected Target Values': interactive plot illustrating the best-so-far functions values as a function of the budget, in particular the averages $\widehat{\mathbb{E}}[V(A,f,t)]$, the median $\widehat{\Q}[V(A,f,t),50]$, and standard deviations, 
		\item `Probability Mass Function': interactive histograms of the $V(A,f,t,i)$ values for budgets $t$ selected by the user and an approximation for an empirical probability mass function for $V(A,f,t)$, and
	\item `Cumulative Distribution': ECDF curves and normalized values for the area under the ECDF curve for budgets specified by the user.
\end{itemize}
The plots are similar to those presented in Section~\ref{sec:post:target}, we omit a detailed description. 

\subsection{Parameter Evolution}
\label{sec:post:params}

In this section the user can track the evolution of the parameters (cf. Section~\ref{sec:example2} for an example explaining how to record this data in the experimental part of \tool). In the example of Figure~\ref{fig:post:target:params}, we see that the Algorithm \emph{LeadingOnes\_1\_plus\_50\_adap\_p} (red curve) used a static population size of 50, while Algorithm \emph{LeadingOnes\_1\_plus\_10\_adap\_lambda} (green curve) uses a dynamic population size. Starting from solutions of function value 80, the average population size of this algorithm was around $163$. A table containing average values as well as quantiles and standard deviations can be downloaded/stored on the bottom of this tab. 

The corresponding fixed-budget results will be made available shortly. 

\begin{figure}
\centering
\includegraphics[width=\linewidth]{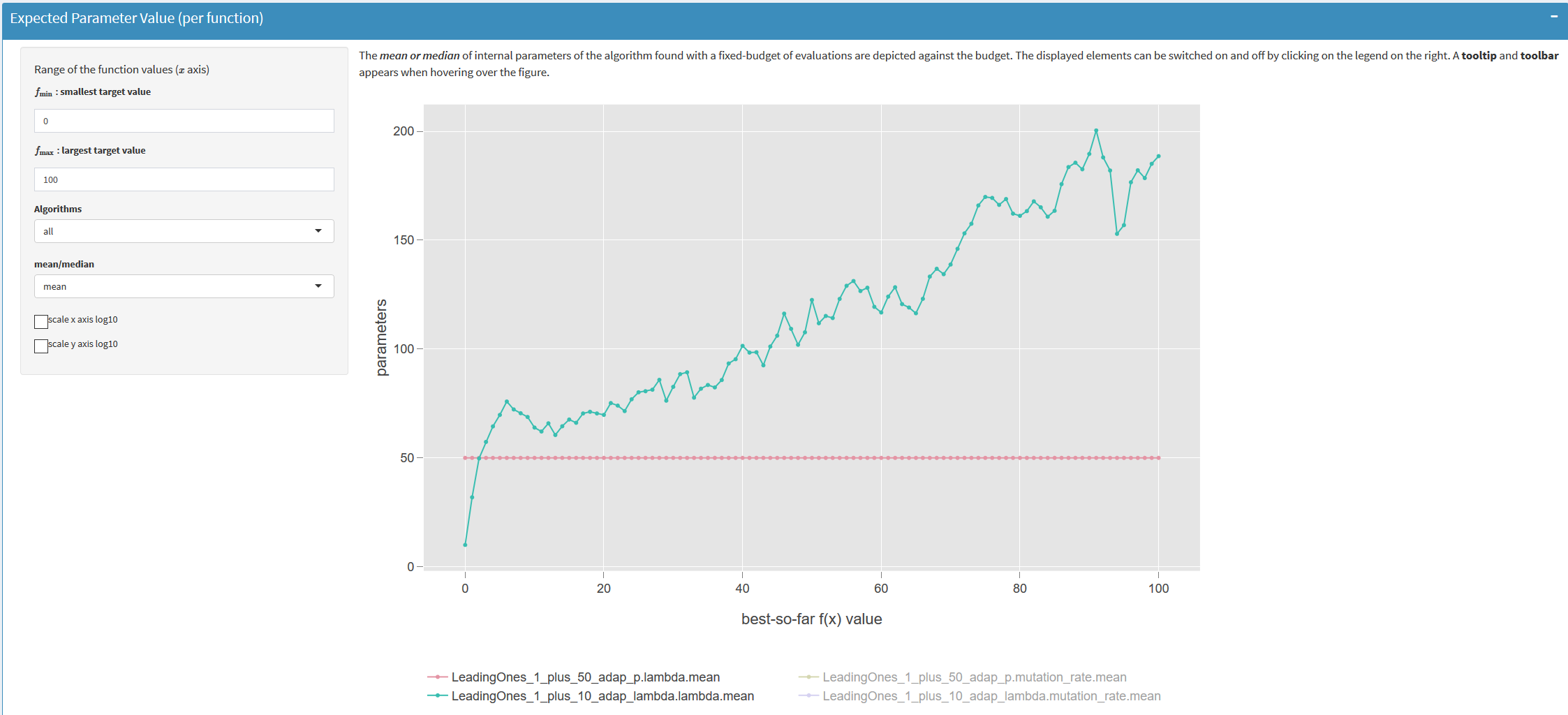}
\caption{`Algorithm Parameters': Average parameter values at given target values.}
\label{fig:post:target:params}
\end{figure}

\section{Conclusions and Possible Extensions}\label{sec:conclusions}

An important aspect of benchmarking, which we have taken aside in this present work, is the \textbf{selection of suitable benchmark problems.} The user can apply \tool to any set of optimization problems that can be formulated as maximization of a function $f:\mathcal{S} \to \R$. In ongoing collaborations with various colleagues, most notably working group 3 from COST action CA15140, we aim to present to the community a suggestion of benchmark functions that should be included in a standardized benchmark set. We recall that the continuous counterpart COCO~\cite{hansen2016cocoplat}, on which \tool is built, compares in the single-objective, static, and noise-free case performance across 24 functions, which are grouped into 5 sets, according to whether or not they are separable, uni- or multi-modal, well- or ill-conditioned, and according to whether or not they exhibit a global structure, cf.~\cite{BBOBfunctions} for details. For the discrete benchmarking, we suggest to start the discussion which problems to include in the benchmark environment by the question which problem features should be represented, and across which of them performance should be aggregated. 

All performance indicators provided by \tool are based on counting function evaluations. In the long run, it will be desirable to allow for a comparison between iterative and non-iterative optimization methods such as Mathematical Programming. To this end, the outputs of \tool will have to be adjusted to \textbf{time-based performance indicators.} A major challenge posed by the latter is the question if or how to provide system-independent performance measures, i.e., results that do not depend on the hardware on which the algorithms are run.

An important aspect that we plan to address in future work is the extension of \tool to allow for comparisons of noisy, dynamic, constrained, or multi-objective optimization problems. Finally, we also consider to extend \tool to other search domains, e.g., permutation-based problems.

As a short-term perspective, we will include additional performance measures, in particular a comparison across different dimensions and some standard statistical tests. Concerning the experimental part, we are most notably working on simplifying the creation of different problem suits.

\subsubsection*{Acknowledgments} 

We thank our colleagues Anne Auger, Dimo Brockhoff, Arina Buzdalova, Maxim Buzdalov, Johann Dr\'eo, Nikolaus Hansen, Pietro S. Oliveto, Ofer Shir, Markus Wagner, and Thomas Weise for various discussions around the benchmarking of iterative optimization heuristics.  

Parts of our work have been inspired by working group 3 of COST Action CA15140 ‘Improving Applicability of Nature-Inspired Optimisation by Joining Theory and Practice (ImAppNIO)’ supported by the European Cooperation in Science and Technology.

Our work has been supported by a public grant as part of the Investissement d'avenir project, reference ANR-11-LABX-0056-LMH, LabEx LMH, in a joint call with the Gaspard Monge Program for optimization, operations research, and their interactions with data sciences.

Furong Ye acknowledges financial support from the China Scholarship Council, CSC No. 201706310143.

}
\newcommand{\etalchar}[1]{$^{#1}$}

\appendix
\section*{Appendix}

\section{Manual for the Experimental Part: Data Generation}
\label{sec:pre}

In this section we describe the experimental part of \tool and provide an illustrated manual, which enables the user to conduct experiments of their choice. We recall that this part is build upon the COCO (COmparing Continuous Optimisers) platform~\cite{hansen2016cocoplat}, from which the data structure and some tool functions are inherited.

\subsection{Preparation}
Running \tool requires a working python environment and a C compiler. The benchmark problems as well as the algorithms can be provided in either python or in C. \tool has currently been tested with python 2.7.12 and with gcc 5.4.1. A version accepting algorithms and problems in Java is in preparation.

Both the experimental and the post-processing parts of \tool can be downloaded from the GitHub page \url{https://github.com/IOHprofiler}. After downloading the zipped files of the experimental part, the archive needs to be extracted.

\subsection{Overview of the Main Steps}

The files of the C [python] interface are located at the path 
"/code-experiments/build/c" ["/code-experiments/build/python"]. 

To run an experiment, the following steps need to be executed. 
\begin{itemize}
    \item \textbf{1. Benchmark Selection and Configuration:} The user needs to select the set of problems (hereafter called "the suite") for which benchmark data shall be generated. The definition of the suite is done in the configuration file \file{configuration.ini}, which is also used to select the granularity at which performance data is stored, the location at which the results are stored, etc. The configuration file and the suite definition are described in Section \ref{sec:Cgf}.
    \item \textbf{2. Algorithm Setup:} The algorithm for which the performance data is generated needs to be defined in the file \file{user\_algorithm.c} [\file{user\_algorithm.py}]. To this end, the content of the function "user\_algorithm" (which includes as example the code for pure random search) is replaced by the new algorithm. 
    \item \textbf{3. Data Generation}: After ensuring that the working path is "code-experiments/build/c" ["code-experiments/build/python"], the execution of the statement "python ../../../do.py run-c" ["python ../../../do.py run-python"] generates the performance data, which is saved in the current path.   
\end{itemize}

\subsection{The Configuration File}
\label{sec:Cgf}

\tool applies the INI file format for the configuration file "configuration.ini". All variables are grouped into three sections: [suite], [observer], and [triggers].\\

The section \textbf{[suite]} contains information about the benchmark problems for which performance data is generated. There are four keys in this section:
\begin{itemize}
    \item suite\_name is the name of the suite. The suite is defined in the file \file{suite\_PBO.c} at "/code-experiments/scr/". As an example, the suite PBO, which contains the benchmark problems \OneMax, \textsc{LeadingOnes}, a linear function with random weights between 0 and 5, and a jump function with gap size $k=1$ is pre-defined as an example. For the time being, we recommend that users keep using the PBO suite, and add their benchmark functions to it. The design of a new suite is possible, but not recommended.
    \item functions\_id: IDs of the selected benchmark problems. An en-dash \textquotesingle  -\textquotesingle is allowed to present the range of problem IDs, for example, "1-4". Alternatively, the problems can be listed by comma, for example, "1,2,3,4". 
    An explanation of how to add new benchmark problems and how to assign the function IDs will be given in Section~\ref{sec:newproblems}. In the distributed version the following benchmark problems are already defined:
    \begin{enumerate}
        \item [1] OneMax
        \item [2] LeadingOnes
        \item [3] Jump with jump size $k=1$ 
        \item [4] A linear function with fixed, but randomly chosen weights between 0 and 5. 
    \end{enumerate}
    \item instances\_id: IDs of the problem instances, cf. Section~\ref{sec:instances}. 
    Instances can be selected using commas and en-dashes, e.g., "1-25,75,80-100".
    \item dimensions: the selection of the problem dimensions, e.g., "100,500,1000" will create experimental data for the three different problem dimensions.
\end{itemize}

The \textbf{[observer]} section contains information concerning the output files. There are five keys in this section:
\begin{itemize}
    \item observer\_name: the name of observer. We suggest to use PBO for the time being, and updated description will be made available when the functionality to create new suits has been improved. 
    \item result\_folder: the name of folder where the results will be stored. If the folder does not exist, it will be created automatically.
    \item algorithm\_name: a name for the algorithm. This information will be stored in the output files.
    \item algorithm\_info: users can write here additional information about the algorithm here, this information will be stored in the .info file of the output, cf. Section~\ref{sec:intro:precision}.
    \item parameters\_name: a list of parameters to be stored. If no algorithm parameters are to be stored, the users leaves this as "". If several parameters are to be stored, they are separated by a comma, e.g., "p1, p2, p3".
\end{itemize}

In the \textbf{[triggers]} section the user decides the granularity of the performance data. According to the user's choice, up to four different files will be created, cf. Section~\ref{sec:intro:precision} for a description of the available output files. 

There are four keys in the [triggers] section.
\begin{itemize}
    \item complete\_triggers: Set as "true" to output *.cdat files.
    
    \item number\_interval\_triggers: The step size for *.idat files. For example, selecting number\_interval\_triggers = 50 will store results of every 50-th function evaluation.
		If the user does not wish to generate  *.idat files, he/she selects number\_interval\_triggers = 0. 
		
    \item number\_target\_triggers: the budget $t$ of storing information for every $[10^i,10^{i+1}]$ in *.tdat files. 
    If the user does not wish to generate  *.tdat files, he/she selects number\_target\_triggers = 0.    
    \item base\_evaluation\_triggers: A set of parameter $v$ for *.tdat files, for example, "1,2,5" means storing information every $1*10^i$-th, $2*10^i$-th and $5*10^i$-th function evaluation in *.tdat files.
     If the user does not wish to generate  *.tdat files, he/she selects base\_evaluation\_triggers = 0.
\end{itemize}

\subsection{Adding New Benchmark Problems}
\label{sec:newproblems}

\tool allows to add new user-defined benchmark problems. To do so, the user needs to create a problem file, which contains the definition of the function, and needs to include the problem into a suite. 

The benchmark problems are defined in \file{f\_*.c} files. The probably easiest way to create a new benchmark problem is to copy the \OneMax example contained in file \file{f\_one\_max.c}, and to adjust it to the new function. 

The actual definition of the \OneMax function is contained in the function f\_one\_max\_raw. The user should replace the content of this function by his/her problem. All occurrences of  "one\_max" need to be replaced by the name of the new problem (we use ``new\_problem'' in the following). If different problem instances are desired, the user creates these in the function f\_new\_problem\_IOHProfiler\_problem\_allocate, cf. Section~\ref{sec:instances} for details.

To be used by \tool, the new problem needs to be added to a problem suite. For example, the pre-installed suite "PBO" is defined in the file \file{suite\_PBO.c}. To include the new problem in this suite, the problem file \file{f\_new\_problem.c} is added to the header of the file \file{suite\_PBO.c}. Then, the f\_new\_problem\_IOHProfiler\_problem\_allocate function of the new problem needs to be called in the function PBO\_get\_problem. Finally, the problem numbers need to be modified in the function suite\_PBO\_initialize.

\subsection{Problem Instances}
\label{sec:instances}

Many standard IOHs are representation-invariant, in the sense that their performance is identical on each fitness landscape, regardless of how it is embedded. That is, the performance is oblivious of rotations and shifts. Furthermore, it is sometimes argued that performance should also be oblivious with respect to a scaling of the function values---algorithms respecting this invariant are often referred to as ``comparison-based''. Users interested in testing how sensitive their algorithms are with respect to search space and/or fitness landscape transformations can make use of build-in transformations: To this end---similar to the COCO framework---\tool offers to test performance on various instances of the same problem. Precisely, the following four transformations are available. They can be combined with each other, cf. also Section~\ref{sec:functions} of the main document. As mentioned above, these transformations are chosen by the user in function f\_new\_problem\_IOHProfiler\_problem\_allocate in the file \file{f\_new\_problem.c}.
\begin{itemize}
    \item \textbf{transform\_obj\_scale:} multiplicative shift of the function values, i.e., instead of $f(x)$ the transformed function values $af(x)$ are returned to the algorithm. 
    \item \textbf{transform\_obj\_shift:} additive shift of the function values, i.e., instead of $f(x)$ the algorithms receive the transformed function values $f(x) + b$. 
    \item \textbf{transform\_vars\_xor:} an XOR of the search point, i.e., a shift in the search space. Instead of evaluating $f(x)$, the function values $f(x\oplus z)$ are computed and returned to the algorithm. Note that in this case the instance $f(\cdot \oplus z)$ has a fitness landscape that is isomorphic to that of the original function $f$. Algorithms that are \emph{unbiased} in the sense of~\cite{LehreW12,ABB,DoerrKLW13} show the same performance on any of these instances.   
    \item \textbf{transform\_vars\_sigma:} a permutation of the search point, i.e., instead of computing function values $f(x_1,x_2,...,x_n)$, the algorithms receive the function values $f(\sigma(x))$ of the permuted search points $\sigma(x)= (x_{\sigma(1)},x_{\sigma(2)},...,x_{\sigma(n)}$, where $\sigma$ is a permutation of the set $\{1,2,\ldots, n\}$.
\end{itemize}

Using \OneMax as an example, we demonstrate how to define the different problem instances. As a general rule, we recommend to reserve instance ``1'' for the original benchmark problem, i.e., the one that does not call any of the four transformations. 

We first demonstrate to convert the original \OneMax function $f$ so that instead of $f(x)$ values the algorithm receives the function values $af(x \oplus z)+b$. To this end, we first assign to "problem" the original \OneMax instance (cf. line~1 in the example below). After that, the instance $f(\cdot)$ is transformed to $f(\cdot \oplus z)$ (line 6), where $z$ is a pseudo-randomized binary vector chosen in line 2. In lines 7 and 8, we then transform the instance $f(\cdot \oplus z)$ to $af(\cdot \oplus z)$ and to $af(\cdot \oplus z)+b$, respectively. The multiplicative and additive shifts "$a$" and "$b$" are again two pseudo-random numbers, which are chosen in lines 3-5. The ranges for $a$ and $b$ are $[1/5,5]$ and $[-1000,1000]$, respectively. 
\begin{lstlisting}
problem = f_one_max_allocate(dimension);
IOHprofiler_compute_xopt(z,rseed,dimension);
a = IOHprofiler_compute_fopt(function,instance + 100);
a = fabs(a) / 1000 * 4.8 + 0.2;
b = IOHprofiler_compute_fopt(function,instance);
problem = transform_vars_xor(problem,z,0);
problem = transform_obj_scale(problem,a);
problem = transform_obj_shift(problem,b);
\end{lstlisting}

The following code shows how to transform the original instance $f(\cdot)$ to $af(\sigma(\cdot))+b$, where we recall that we denote by $\sigma(x)$ the permuted string $(x_{\sigma(1)}, \ldots, x_{\sigma(n)})$. 
After being allocated with the original \OneMax instance $f(\cdot)$, "problem" is transformed to $f(\sigma(\cdot))$ in line~13, where $\sigma$ is a pseudo-random permutation chosen in lines~3 to~9. Then, following the same procedure as in the example above, "problem" is transformed to $af(\sigma(\cdot))$ in line~14 and, finally, to $af(\sigma(x))+b$ in line~15, where $a$ and $b$ are again random numbers, chosen in lines~10 to~12.
\begin{lstlisting}
problem = f_one_max_allocate(dimension);
IOHProfiler_compute_xopt_double(xins,rseed,dimension);
for(i = 0; i < dimension; i++){
	sigma[i] = i;
}
for(i = 0; i < dimension; i++){
	t = (int)(xins[i] * dimension);
	temp = sigma[0];sigma[0] = sigma[t];sigma[t] = temp; 
}
a = IOHProfiler_compute_fopt(function,instance + 100);
a = fabs(a) / 1000 * 4.8 + 0.2;
b = IOHProfiler_compute_fopt(function, instance);
problem = transform_vars_sigma(problem, sigma, 0);
problem = transform_obj_scale(problem,a);
problem = transform_obj_shift(problem,b);
\end{lstlisting}

\subsection{Examples}
\label{sec:examples}

The user can find two examples in the git folder \file{/example/}:
\begin{itemize}
    \item \file{/example/example1/} includes the code and the results of pure random search (hereafter called "random search"), while
    \item \file{example/example2/} includes the code and results of a $(1+\lambda)$ evolutionary algorithm.
\end{itemize}
We describe the configuration of these examples in Sections~\ref{sec:example1} and~\ref{sec:example2}. 

\subsubsection{Example 1: Pure Random Search}\label{sec:example1}
The example of the "random search" method is located at the path \file{/example/example1/}. 

The "configuration.ini" file is set as follows.

\noindent
\textbf{$[$suite$]$}\\
suite\_name = PBO\\
functions\_id = 1-4\\
instances\_id = 1-100\\
dimensions = 100\\
\textbf{$[$observer$]$}\\
observer\_name = PBO\\
result\_folder = EXP\\
algorithm\_name = RANDOM\_SEARCH\\
algorithm\_info = RANDOM\_SERACH\\
parameters\_name = evaluation\\
\textbf{$[$triggers$]$}\\
complete\_triggers = true\\
number\_interval\_triggers = 10\\
number\_target\_triggers = 3\\
base\_evaluation\_triggers = 1,2,5

Based on this configuration file, performance data is collected for the algorithm optimizing the 100-dimensional variants of the benchmark problems 1, 2, 3, and 4. For each problem, the instances from 1 to 100 are used. For each instance the number of independent runs performed in the experimental part is specified in the variable "INDEPENDENT\_RESTARTS" in the file containing the algorithm, cf. example below. If, for example, the user wishes to run each of the instances 1-100 instances twice, he/she sets "instance\_id = 1-100" in the configuration file, and sets "INDEPENDENT\_RESTARTS = 2" in the algorithm file. 

The results of this example experiment will be stored in the folder \file{./EXP/}.  The name of the parameter to be stored is set as "evaluation"; this will be the header of the respective column in the output files.

In this example four different output files will be created: \\
- *.cdat storing data of each iteration, \\
- *.idat storing data from every 10-th iteration,\\
- *.tdat storing data from every $10^{n/3}$-th, every $1*10^n$-th, $2*10^n$-th, and $5*10^n$-th function evaluation, \\
- *.dat storing data for each iteration in which an improvement has been found,
where * is of the form IOHprofiler\_f1\_DIM100\_i1, as explained in Section~\ref{sec:intro:precision}.

The user\_algorithm that implements the pure random search is implemented in file \file{/example/example1/c/user\_algorithm.c} as follows: 

\begin{lstlisting}
static const size_t BUDGET_MULTIPLIER = 50;
static const size_t INDEPENDENT_RESTARTS = 1;
void User_Algorithm() {
  size_t number_of_parameters = 1;
  int *x = IOHProfiler_allocate_int_vector(dimension);
  double *y = IOHProfiler_allocate_vector(number_of_objectives);
  double *p = IOHProfiler_allocate_vector(number_of_parameters);
  size_t i, j;

  for (i = 0; i < max_budget; ++i) {
    for (j = 0; j < dimension; ++j) {
      x[j] = (int)(IOHProfiler_random_uniform(random_generator) * 2);
    }
    p[0] = i + 1;
    set_parameters(number_of_parameters,p);
    evaluate(x, y);
  }

  IOHProfiler_free_memory(x);
  IOHProfiler_free_memory(y);
}
\end{lstlisting}

The user needs to set two parameters in the \file{user\_algorithm.c} [\file{user\_algorithm.py}].
\begin{itemize}
    \item BUDGET\_MULTIPLIER: This parameter controls the maximal budget of function evaluations, which is set to BUDGET\_MULTIPLIER times the dimension of the problem.
    \item INDEPENDENT\_RESTARTS: The number of independent runs of the algorithm for each instance. That is, setting "INDEPENDENT\_RESTARTS=100" and "instance\_id=1-3" will result in an overall number of 300 runs---100 independent runs for each of the first three instances.
\end{itemize}

With the code above, the maximal number of evaluations for each run is 50*dimension, and the algorithm will not restart within one run.

For each iteration, a new individual x is generated randomly (line 12), and the fitness is evaluated by line 16. Note here that y is a vector that stores the fitness of x. Also, a parameter p (evaluation step) will be logged in output files (line 15), and its logging name is defined in the configuration file as "evaluation". We will see in the next section an example where more than one parameter are stored. 

\subsubsection{Example 2: A \texorpdfstring{($1+\lambda$)}{(1+lambda)} EA }
\label{sec:example2}
This example of a ($1+\lambda$) EA is located at the path \file{/example/example2/}. 
The configuration file is as follows.

\noindent
\textbf{$[$suite$]$}\\
suite\_name = PBO\\
functions\_id = 1-4\\
instances\_id = 1\\
dimensions = 100,500,1000\\
\textbf{$[$observer$]$}\\
observer\_name = PBO\\
result\_folder = EXP\\
algorithm\_name = ONE\_PLUS\_LAMDA\_EA\\
algorithm\_info = ONE\_PLUS\_LAMDA\_EA\\
parameters\_name = mutation\_rate,l\\
\textbf{$[$triggers$]$}\\
complete\_triggers = true\\
number\_interval\_triggers = 50\\
number\_target\_triggers = 0\\
base\_evaluation\_triggers = 0\\

Based on this configuration file, running time data is generated for the benchmark problems with function IDs 1 to 4. For each function, the algorithm will be run for dimension 100, 500, and 1000. Only the first instance (without transformation) is tested. All results will be stored in the folder \file{./EXP/}, and the names of the two parameters which are tracked  are set as "mutation\_rate" and "l" (the number of bits in which parent and offspring differ).

The three data files *.idat (storing data after every 50-th evaluation), *.cdat, and *.dat will be generated. 

The user\_algorithm that implements the ($1+\lambda$) EA is as follows:

\begin{lstlisting}[language=Python]
import numpy as np
import random as rd
import math

independent_restart = 10
budget = 50 

def mutation(ind,mutation_rate,dim):
    l = 0
    while l == 0:
        l = np.random.binomial(dim,mutation_rate)
    flip = rd.sample(range(0,dim),l)

    for index in flip:
        ind[index] = (ind[index] + 1) % 2

    return l

def user_algorithm(fun,lbounds,ubounds,budget):
    lbounds, ubounds = np.array(lbounds), np.array(ubounds)
    dim = fun.dimension
    parent = lbounds + (ubounds - lbounds + 1) * np.random.rand(dim)
    parent = parent.astype(int)
    best = parent.copy()
    budget -= 1
    lamb = 1
    mutation_rate = 1.0/dim
    para = np.array([mutation_rate])
    fun.set_parameters(para)
    best_value = fun(parent)
    while budget > 0:
        for i in range(0,lamb):
            offspring = parent.copy()
            l = mutation(offspring,mutation_rate,dim)
            para = np.array([mutation_rate,l])
            fun.set_parameters(para)
            v = fun(offspring)
            if v > best_value :
                best_value = v
                best = offspring.copy()
            budget -= 1
            if(budget == 0):
                break
        parent = best.copy()
        mutation_rate = 1.0 / (1 + (1 - mutation_rate) / mutation_rate * math.exp(0.22 * np.random.normal()))
        mutation_rate = min(max(mutation_rate,1.0/dim),0.5)
    return best_value
\end{lstlisting}

With the code above, the maximal number of function evaluations for each run is 50*dimension, and the algorithm will do ten independent runs for each selected instance (as discussed above, only instance 1 has been selected in the configuration file).

For each generation, $\lambda$ offspring are created by mutating the parent individual (line 34). Their fitness is evaluated in line 36. The function fun(offspring) returns the fitness of the offspring.

In addition to the information about the fitness values, a vector of parameters ('para') will be stored in the output files (line 36). The vector stores the mutation\_rate and the number of flipped bits (line 35). The names attributed to these parameters are chosen as "mutation\_rate, l" in the \file{configuration.ini} file. 

\subsection{Overview of the Different Files}
\label{app:files}

The following lists summarize the folders and the main files that can be found in the experimental part of \tool. The files that need to be edited by the user are formatted in bold font. 

Folder \textbf{/code-experiments/}:
\begin{itemize}
\item /src/ : a folder of source files
\item /build/ : a folder of C and Python interfaces
\item /tools/ : some common tools for the project
\end{itemize}
Folder \textbf{/example/}:
\begin{itemize}
\item /example1/ : examples of random search method, cf. Section~\ref{sec:example1}
\item /example2/ : examples of ($1+\lambda$) EA, cf. Section~\ref{sec:example2}
\end{itemize}
Folder \textbf{/src/}:
\begin{itemize}
\item f\_binary.c : Implementation of the binary function and problem
\item f\_jump.c : Implementation of the jump function and problem
\item f\_leading\_ones.c : Implementation of the leading ones function and problem
\item f\_linear.c : Implementation of the linear function and problem
\item f\_one\_max.c : Implementation of the one\_max function and problem
\item IOHProfiler.h : Header file for all public IOHProfiler functions and variables
\item IOHProfiler\_internal.h : Definitions of internal IOHProfiler structures and typedefs
\item IOHProfiler\_observer.c : Definitions of functions regarding IOHProfiler observers
\item IOHProfiler\_platform.h : Automatic platform-dependent configuration of the IOHProfiler framework
\item IOHProfiler\_problem.c : Definitions of functions regarding IOHProfiler problems
\item IOHProfiler\_random.c : Definitions of functions regarding IOHProfiler random numbers
\item IOHProfiler\_runtime\_c.c : Generic IOHProfiler runtime implementation for the C language
\item IOHProfiler\_string.c : Definitions of functions that manipulate strings
\item IOHProfiler\_suite.c : Definitions of functions regarding IOHProfiler suites
\item IOHProfiler\_utilities.c : Definitions of miscellaneous functions used throughout the IOHProfiler framework
\item suite\_PBO\_legacy\_code.c : Methods for generating pseudo random numbers
\item logger\_PBO.c : Implementation of the PBO logger
\item observer\_PBO.c : Implementation of the PBO observer
\item \textbf{suite\_PBO.c} : Selection of functions to be included in the PBO suite 
\item transform\_obj\_shift.c : Implementation of shifting the objective value by the given offset
\item transform\_obj\_scale.c : Implementation of scaling the objective value by the given offset
\item transform\_vars\_shift.c : Implementation of shifting all decision values by an offset
\item transform\_vars\_xor.c : Implementation of the xor of all decision values by an offset
\item transform\_vars\_sigma.c : Implementation of re-ordering all decision values by permuting the string of decision values
\end{itemize}
\
Folder \textbf{/build/c/}:
\begin{itemize}
\item Makefile : Makefile to build the C program
\item user\_experiment.c : The interface to invoke user algorithm
\item \textbf{user\_algorithm.c} : The file where the user defines his/her algorithm
\item \textbf{configuration.ini} : the configuration file, cf. Section~\ref{sec:Cgf}
\item ...
\end{itemize}
\
Folder \textbf{/build/python/}:
\begin{itemize}
\item user\_experiment.py : The interface to invoke user algorithm 
\item \textbf{user\_algorithm.py} : The file where the user defines his/her algorithm
\item \textbf{configuration.ini} : the configuration file, cf. Section~\ref{sec:Cgf}
\item ...
\end{itemize}


\end{document}